\documentclass[10pt,twocolumn,letterpaper]{article}

\usepackage[pagenumbers]{cvpr} 

\usepackage[dvipsnames]{xcolor}

\definecolor{cvprblue}{rgb}{0.21,0.49,0.74}
\usepackage[pagebackref,breaklinks,colorlinks,citecolor=cvprblue]{hyperref}

\usepackage{bm}
\usepackage{graphicx}
\usepackage{float}
\usepackage{wrapfig}
\usepackage{enumitem} 
\usepackage{graphicx}
\usepackage{wrapfig}
\usepackage{algorithm}
\usepackage{setspace}
\usepackage{multicol}
\usepackage{multirow}
\usepackage{colortbl}
\usepackage{makecell}
\usepackage{wrapfig} 
\usepackage{pifont} 
\usepackage[accsupp]{axessibility}

\usepackage{amssymb}

\definecolor{cred}{HTML}{FF6B6B}
\definecolor{cyellow}{HTML}{FEC260}
\definecolor{cgreen}{HTML}{70AD47}
\definecolor{cblue}{HTML}{4D96FF}
\definecolor{cpurple}{HTML}{2A0944}
\definecolor{ggray}{HTML}{F5F7F8}
\definecolor{ggray}{HTML}{EEEEEE}
\definecolor{aliceblue}{rgb}{0.94, 0.97, 1.0}

\makeatletter
\newcommand{\ssymbol}[1]{$^{\@fnsymbol{#1}}$}
\makeatother

\newcommand{\myparagraph}[1]{\textbf{#1}\hspace{1.8ex}}

\newcommand{\Frst}[1]{{\textbf{#1}}}

\newcommand{\largemodel}[1]{\color{gray}{#1}}

\def\paperID{***} 
\def\confName{CVPR}
\def\confYear{2024}

\title{Chat-UniVi: Unified Visual Representation Empowers Large Language Models with Image and Video Understanding}

\author{%
    Peng Jin$^{1,2,3}$ \quad
    Ryuichi Takanobu \quad
    Wancai Zhang$^{4}$ \quad
    Xiaochun Cao$^{5}$ \quad
    Li Yuan$^{1,2,3}$\footnotemark[1]\\[3pt]
    \small{$^1$School of Electronic and Computer Engineering, Peking University, Shenzhen, China} \quad \small{$^2$Peng Cheng Laboratory, Shenzhen, China} \\
    \small{$^3$AI for Science (AI4S)-Preferred Program, Peking University Shenzhen Graduate School, Shenzhen, China} \\
    \small{$^4$Nari Technology Co.,Ltd., China} \
    \small{$^5$School of Cyber Science and Tech., Shenzhen Campus of Sun Yat-sen University, Shenzhen, China}\\
    \small{jp21@stu.pku.edu.cn} \quad \small{yuanli-ece@pku.edu.cn}
    \\[5pt]
    \href{https://github.com/PKU-YuanGroup/Chat-UniVi}{https://github.com/PKU-YuanGroup/Chat-UniVi}
    }

\begin{document}
\maketitle

\renewcommand{\thefootnote}{\fnsymbol{footnote}}
\footnotetext[1]{Corresponding author: Li Yuan.}

\begin{abstract}
Large language models have demonstrated impressive universal capabilities across a wide range of open-ended tasks and have extended their utility to encompass multimodal conversations. However, existing methods encounter challenges in effectively handling both image and video understanding, particularly with limited visual tokens. In this work, we introduce Chat-UniVi, a \textbf{Uni}fied \textbf{Vi}sion-language model capable of comprehending and engaging in conversations involving images and videos through a unified visual representation. Specifically, we employ a set of dynamic visual tokens to uniformly represent images and videos. This representation framework empowers the model to efficiently utilize a limited number of visual tokens to simultaneously capture the spatial details necessary for images and the comprehensive temporal relationship required for videos. Moreover, we leverage a multi-scale representation, enabling the model to perceive both high-level semantic concepts and low-level visual details. Notably, Chat-UniVi is trained on a mixed dataset containing both images and videos, allowing direct application to tasks involving both mediums without requiring any modifications. Extensive experimental results demonstrate that Chat-UniVi consistently outperforms even existing methods exclusively designed for either images or videos. Code is available at \href{https://github.com/PKU-YuanGroup/Chat-UniVi}{https://github.com/PKU-YuanGroup/Chat-UniVi}.
\end{abstract}

\section{Introduction}
Large language models (LLMs), such as GPT-3~\cite{brown2020language} and LLaMA~\cite{touvron2023llama,touvron2023llama2}, showcase substantial universal capabilities that pave the way for achieving general artificial intelligence. However, language represents just one facet of communication. Visual information serves to augment and enhance our comprehension of the world. Therefore, there exists a burgeoning interest in developing a multimodal conversation model that can accommodate various input modalities simultaneously, including images and videos. 

\begin{figure}[tbp]
\centering
\includegraphics[width=1\linewidth]{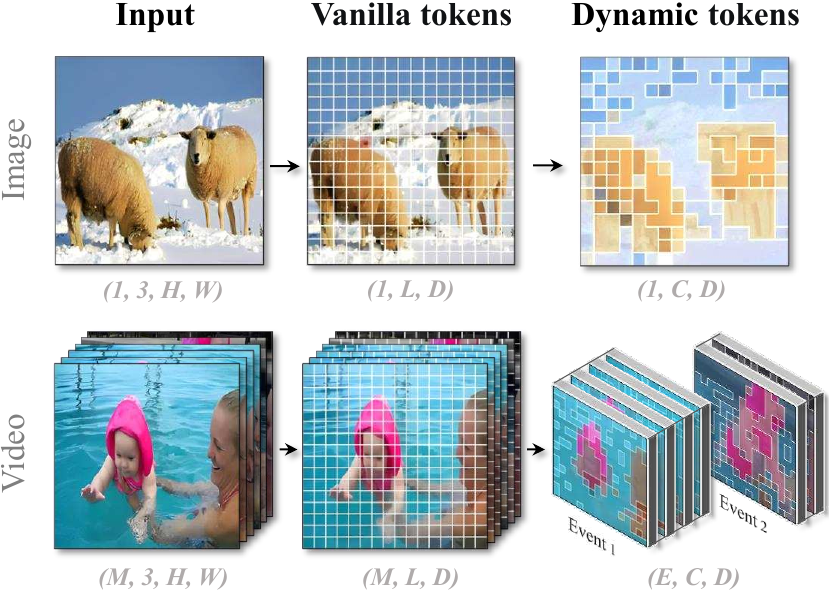}
\vspace{-1.5em}
\caption{\textbf{The unified representation framework for images and videos utilizing a collection of dynamic visual tokens.} ``$H$'' and ``$W$'' represent the height and width of the input, respectively. ``$L$'', ``$D$'', ``$M$'', ``$C$'', and ``$E$'' denote the number of vanilla visual tokens, the feature dimension, the frame length, the number of dynamic visual tokens, and the number of events, respectively.}
\vspace{-.6em}
\label{fig1}
\end{figure}

Recent advances in multimodal conversation models, such as MiniGPT-4~\cite{zhu2023minigpt}, LLaVA~\cite{liu2023visual,liu2023improved}, and mPLUG-Owl~\cite{ye2023mplug}, focus on integrating visual tokens into LLMs. Despite their commendable progress, existing methods often specialize in either image or video inputs. For instance, methods~\cite{liu2023visual,liu2023improved} that prioritize image inputs typically employ a larger number of visual tokens to attain finer spatial understanding. Conversely, methods~\cite{maaz2023video} concentrating on video inputs often compromise spatial comprehension per frame to accommodate more frames for modeling temporal relationships. Although some methods, \eg, Flamingo~\cite{alayrac2022flamingo}, can extract a fixed number of tokens for each image and video using a query transformer, their primary emphasis remains on image understanding, lacking the capability to effectively model temporal comprehension, thus resulting in a limited understanding of videos. Therefore, it is crucial and challenging to enable LLMs for both image and video comprehension within a unified framework.

In this paper, we introduce Chat-UniVi, a \textbf{Uni}fied \textbf{Vi}sion-language model designed to proficiently comprehend and engage in conversations about both images and videos. Chat-UniVi uniformly represents images and videos using a collection of dynamic visual tokens, enabling it to concurrently capture the spatial details of images and the comprehensive temporal relationship of videos. As illustrated in \cref{fig1}, images can be depicted through visual tokens of diverse sizes. For example, the primary object, \ie, the sheep in \cref{fig1}, necessitates a fine-grained representation with numerous visual tokens, while the background, \ie, the snow-capped mountain, can be sufficiently modeled with only one visual token. In the case of videos, the video is initially divided into several events, and subsequently, these visual tokens expand over frames within each event to encapsulate frame-level dynamics. Such unified representation for both images and videos significantly reduces the number of visual tokens while maintaining the expressive capabilities of the model. It is worth noting that longer videos are assigned more visual tokens in our method. Therefore, our method is better suited for variable-length video understanding than existing methods.

To obtain these dynamic visual tokens, we propose a token merging method for progressively merging visual tokens with similar semantic meanings. Specifically, starting with visual tokens initialized by the vision transformer~\cite{dosovitskiy2021an}, we gradually group them by applying the k-nearest-neighbor based density peaks clustering algorithm, \ie, DPC-KNN~\cite{du2016study}, on the token features. When it comes to videos, we also utilize DPC-KNN on the frame features to get events. At each merging step, visual tokens assigned to the same cluster are merged by averaging their token features. Finally, we provide a multi-scale representation to the LLMs, where the upper layers of the multi-scale representation encompass high-level semantic concepts, while the lower layers emphasize visual details representations.

\begin{figure}[tbp]
\centering
\includegraphics[width=1\linewidth]{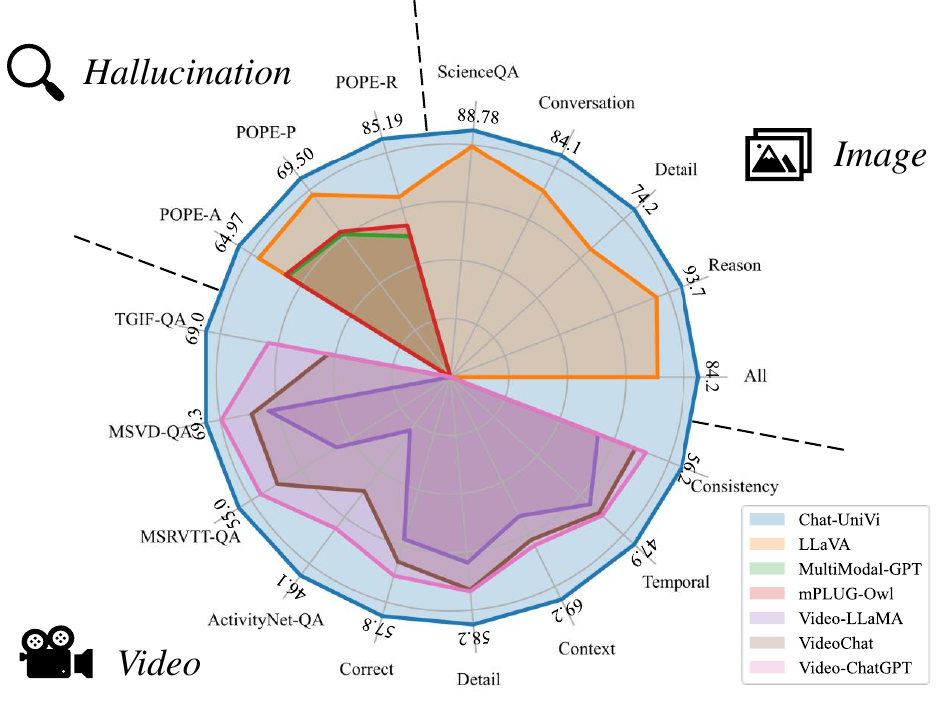}
\vspace{-1.6em}
\caption{\textbf{The proposed Chat-UniVi, designed as a unified model, consistently outperforms even existing methods exclusively designed for either images or videos.} These results demonstrate the advantages of the proposed method.}
\vspace{-.6em}
\label{fig1_1}
\end{figure}

The proposed Chat-UniVi has two compelling advantages: \textbf{First}, its unified image and video modeling method allows training on the mixed dataset of image and video, enabling direct application to both image and video tasks without any modifications. \textbf{Second}, the multi-scale representation contributes to the comprehensive understanding of images and videos, empowering Chat-UniVi to adapt to various tasks, including employing high-level representation for semantic understanding and low-level representation for generating detailed descriptions. We evaluate Chat-UniVi on both image and video understanding tasks. As shown in \cref{fig1_1}, compared to other methods focused exclusively on either images or videos, Chat-UniVi consistently demonstrates superiority in comprehending images and videos. Moreover, we also provide evidence of the advantages of joint training of images and videos for multimodal large language models. The main contributions are summarized as follows:
\begin{itemize}
    \item We propose a unified visual representation for LLMs, enabling LLMs to comprehend both images and videos.
    \item We uniformly represent images and videos using multi-scale dynamic visual tokens and propose a token merging method to obtain these dynamic visual tokens. 
    \item Without fine-tuning, Chat-UniVi attains competitive performance in both image and video tasks and achieves impressive results in the object hallucination benchmark.
\end{itemize}

\begin{figure*}[tbp]
\centering
\includegraphics[width=1\textwidth]{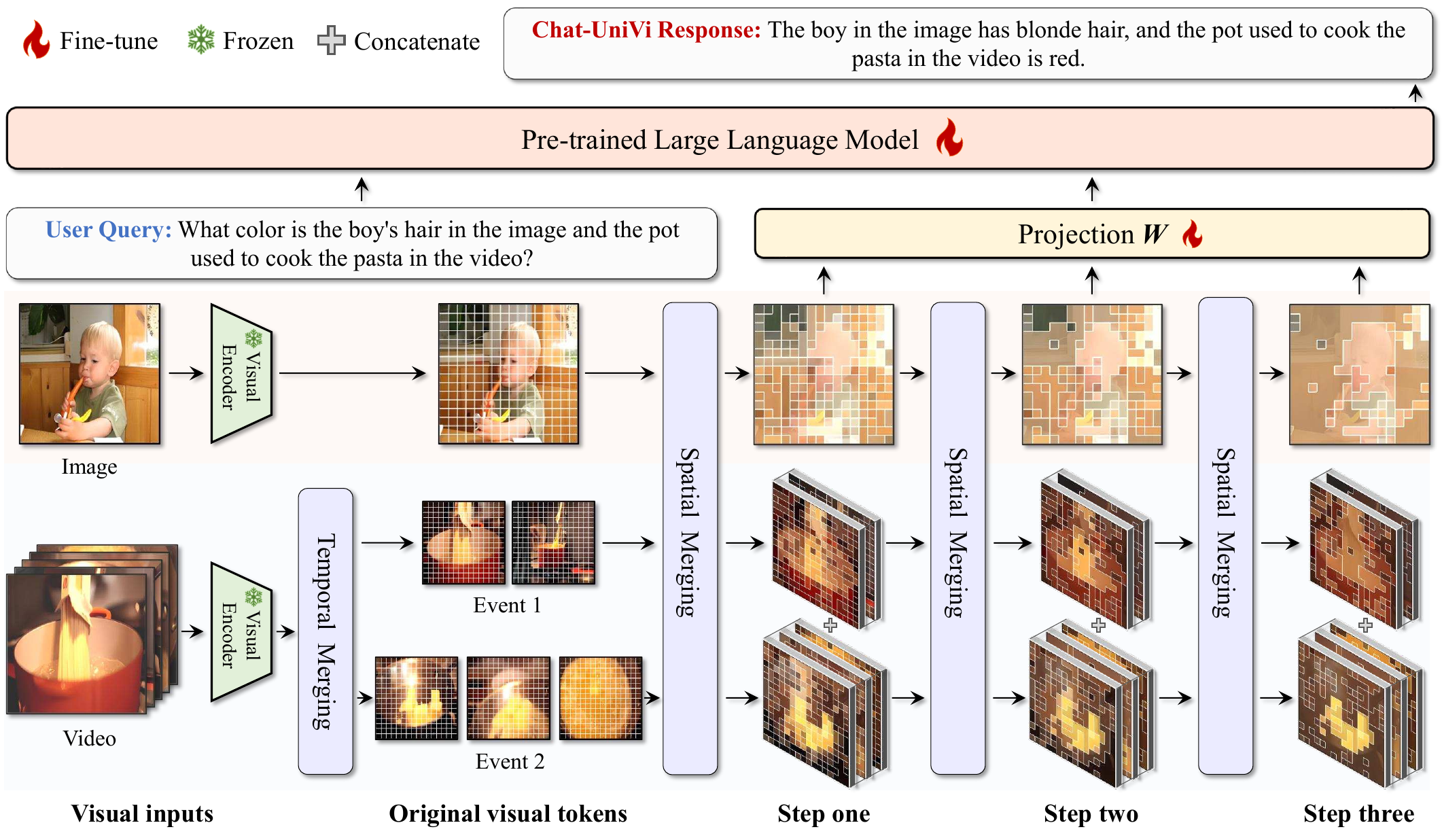}
\vspace{-1.6em}
\caption{\textbf{The overview of the proposed Chat-UniVi for conversations containing both images and videos.} Chat-UniVi uniformly represents images and videos using a collection of dynamic visual tokens and provides a multi-scale representation that equips large language models to perceive both high-level semantic concepts and low-level visual details.}
\vspace{-.6em}
\label{fig2}
\end{figure*}

\section{Related Work}
\noindent \myparagraph{Large Language Models.} 
Large language models~\cite{radford2019language,raffel2020exploring,vaswani2017attention} have made disruptive progress, primarily attributed to the expansion of training data and the substantial increase in model parameters. Inspired by the success of GPT-3~\cite{brown2020language}, numerous LLMs have subsequently been developed, including PaLM~\cite{chowdhery2022palm}, OPT~\cite{zhang2022opt}, BLOOM~\cite{scao2022bloom}, InstructGPT~\cite{ouyang2022training}, and ChatGPT~\cite{chatgpt}. However, language represents just one facet of communication. Visual information serves to augment and enhance our comprehension of the world~\cite{labiosavisual,jin2022expectation,jin2023diffusionret,jin2023video,ijcai2023p0104,bain2021frozen,wang2022omnivl,zhu2023languagebind}. In this work, we introduce Chat-UniVi, designed to comprehend both image and video inputs.

\noindent \myparagraph{Large-scale Multimodal Models.} 
Existing large-scale multimodal models~\cite{bai2023qwen,chen2023shikra,wu2023next,zheng2023minigpt,chen2023minigpt,gao2023llama,gong2023multimodal,chen2023x,li2022blip,luo2023cheap,liu2023aligning,liu2023hallusionbench,zhu2024llmbind,lin2024moe,gao2024sphinx,ren2023timechat} can be broadly categorized into two classes. The first class of methods~\cite{wu2023visual,shen2023hugginggpt,yang2023mm,suris2023vipergpt} involves using LLMs as a dispatch scheduler, facilitating connections between various expert models to handle different vision tasks. The second class of methods~\cite{gpt4,li2023blip,li2023blip} emphasizes the integration of models from different modalities into end-to-end trainable models. More recently, there have also been several dedicated multimodal models tailored for video processing, such as Video-LLaVA~\cite{lin2023video}, Video-ChatGPT~\cite{maaz2023video}, VideoChat~\cite{li2023videochat}, and Video-LLaMA~\cite{zhang2023video}. Despite their commendable progress, existing methods often focus exclusively on either image or video inputs. In this work, we focus on developing an end-to-end trained multimodal model for both image and video tasks. Although Flamingo also supports both image and video inputs, it can only extract a fixed number of tokens for videos of varying lengths with a query transformer. Recent works~\cite{wu2023next,chen2023x} have explored the use of separately pre-trained image and video encoders for processing, but these methods introduce model redundancy and prove challenging to train together. Hence, it does not align with our focus on achieving a unified vision-language model. In contrast to the previous works, the proposed method uniformly represents images and videos using multi-scale dynamic visual tokens.

\noindent \myparagraph{Dynamic Visual Token.} 
There have also been recent methods~\cite{ma2023image,xu2022groupvit,zeng2022not,rao2021dynamicvit,bolya2022token,ren2023testa} to explore the role of dynamic tokens within the transformer framework. However, none of these methods can be directly extended to video. We summarize the advantages of our method as follows: (i)~\textbf{Supporting video input.} In contrast to other methods, Chat-UniVi extends the dynamic token method to incorporate video inputs, achieving the integration of image and video representations for the first time. Our work is the first to demonstrate that this unified representation can reconcile the intricate spatial details of images with the broader temporal understanding required for videos. (ii)~\textbf{Without parameters.} Our clustering method is parameter-free. Interestingly, we find that this parameter-free clustering method serves as the linchpin to the success of our model. We attribute this phenomenon to the gradient instability in multimodal conversation training, which hinders the convergence of parameterized methods. Comparisons of Chat-UniVi and other dynamic token methods are provided in the appendix.

\section{Methodology}
Chat-UniVi aims to model images and videos concurrently within a language sequence that can be comprehended by Large Language Models (LLMs) in a unified framework. Chat-UniVi achieves this by uniformly representing images and videos through a set of dynamic visual tokens, bridging the intricate spatial details of images with the broader temporal comprehension needed for videos. The overview of the proposed Chat-UniVi is shown in \cref{fig2}.

\subsection{Dynamic Visual Tokens for Image and Video}
Building upon the vanilla Vision Transformer, most methods generate equally important visual tokens by dividing the image into regular and fixed grids. However, it is evident that not all regions hold equal significance in vision-language tasks. For example, capturing the background may require only a single visual token. Drawing inspiration from this insight, We amalgamate non-essential tokens to derive dynamic vision regions as input for LLMs.

\noindent \myparagraph{Spatial Visual Token Merging.} 
For an input image, we adopt the vision encoder of CLIP~\cite{radford2021learning} to provide the original visual tokens $\bm{Z}=\{z_i\}_{i=1}^{L}$, where $L$ is the number of visual tokens each image is divided into. To amalgamate non-essential visual tokens, we utilize DPC-KNN~\cite{du2016study}, a k-nearest-neighbor based density peaks clustering algorithm, to cluster the visual tokens. Starting with visual tokens $\bm{Z}=\{z_i\}_{i=1}^{L}$ initialized by the vision transformer, we first compute the local density $\rho_i$ of each token $z_{i}$ according to its $K$-nearest neighbors, which is formulated as:
\begin{equation}
\rho_i=\textrm{exp}\big(-\frac{1}{K}\sum_{z_{k}\in \textrm{KNN}(z_{i}, \bm{Z})}\Vert z_{k}-z_{i} \Vert^2\big),
\label{eq:1}
\end{equation}
where $\textrm{KNN}(z_{i}, \bm{Z})$ is the $K$-nearest neighbors of $z_{i}$ in $\bm{Z} \backslash \{z_{i}\}$. ``$Z \backslash \{z_{i}\}$'' denotes removing $\{z_{i}\}$ from $\bm{Z}$. Intuitively, $\rho_i$ denotes the local density of token $z_{i}$. Then, we compute the distance index $\delta_i$ of the token $z_{i}$:
\begin{equation}
\delta_i=
\begin{cases}
\underset{j:\rho_j>\rho_i}{\textrm{min}} \Vert z_{j}-z_{i} \Vert^2, & \text{if\ $\exists j$\ s.t.\ $\rho_j>\rho_i$.}\\
\ \ \underset{j}{\textrm{max}} \ \ \Vert z_{j}-z_{i} \Vert^2, & \text{otherwise.}
\end{cases}
\label{eq:2}
\end{equation}
In essence, $\delta_i$ represents the distance between the given token $z_{i}$ from other high-density tokens. We identify those tokens with relatively high $\rho_i \times \delta_i$ as cluster centers and then allocate other tokens to their nearest cluster center according to the Euclidean distances. Finally, we utilize the average token within each cluster to represent the corresponding cluster. The vision region of the merged token is the union of the vision regions within the corresponding cluster.

\noindent \myparagraph{Temporal Visual Token Merging.} 
To adapt the dynamic visual tokens to video inputs, we extend the visual tokens across frames. However, directly consolidating all frames into a limited number of visual tokens may lead to the loss of temporal information within the video. For example, in \cref{fig2}, the video demonstrates the process of cooking pasta before preparing the sauce. Simply merging all frames would pose challenges for the model in determining the correct sequence, such as whether to prepare the sauce first, cook the pasta first, or simultaneously cook the pasta while preparing the sauce. Therefore, we propose temporal visual token merging to first divide the video into several critical events. Subsequently, we make the visual tokens only expand over frames within the same event.

Given the $m_{th}$ frame $\bm{Z}^m=\{z_i^m\}_{i=1}^{L}$ of a video, we first apply mean-pooling over all tokens to obtain the frame-level representation $f^m$. Similar to the spatial visual token merging method, we leverage DPC-KNN to amalgamate non-essential frames. Specifically, we first compute the local density $\rho^m$ and the distance index $\delta^m$ of each frame $f^{m}$. Frames with relatively high $\rho^m \times \delta^m$ are identified as cluster centers, and other frames are then assigned to their nearest cluster center based on Euclidean distances. We treat each cluster as a critical event and denote the set of indexes of the frames in the cluster as $\bm{F}$. Therefore, the set of visual tokens within the $n_{th}$ event $\bm{F}_{n}$ can be formulated as:
\begin{equation}
\tilde{\bm{Z}}_{n}=\big\{z_i^m| m \in \bm{F}_n,\ i \in \{1,2,...,L\}\big\}.
\end{equation}
After completing the temporal visual token merging, we obtain the set of visual tokens within the event, \ie, $\tilde{\bm{Z}}$. To make the visual tokens expand over frames within the event, we adjust \cref{eq:1} and \cref{eq:2} in the spatial visual token merging method to the following form:
\begin{equation}
\begin{aligned}
\tilde{\rho_i}&=\textrm{exp}\big(-\frac{1}{K}\sum_{z_{k} \in \textrm{KNN}(z_{i}, \tilde{\bm{Z}})}\Vert z_{k}-z_{i} \Vert^2\big),\\
\tilde{\delta_i}=&
\begin{cases}
\underset{j:\tilde{\rho_j}>\tilde{\rho_i}}{\textrm{min}} \Vert z_{j}-z_{i} \Vert^2, & \text{if\ $\exists j$\ s.t.\ $\tilde{\rho_j}>\tilde{\rho_i}$.}\\
\ \ \underset{j}{\textrm{max}} \ \ \Vert z_{j}-z_{i} \Vert^2, & \text{otherwise.}
\end{cases}
\end{aligned}
\end{equation}

Finally, we concatenate the expanded dynamic visual tokens together in order of events to ensure the broader temporal understanding required for videos.

\noindent \myparagraph{Multi-scale Representation.} 
To further enhance the capabilities of our model, we propose a multi-step aggregation method designed to provide multi-scale visual features for LLMs. Specifically, in Chat-UniVi, the initial visual tokens at the first merging step are derived from the vision encoder of CLIP. Then, we progressively merge visual tokens with similar semantic meanings and obtain different numbers of tokens in different steps. The higher-level features encompass abstract semantic concepts, while the lower levels emphasize representations of visual details. In practice, we execute a three-step aggregation process for each input image or video. Finally, we concatenate the outputs from each merging step and utilize a trainable projection matrix $\bm{W}$ to transform these multi-scale visual features into language embedding tokens, which serve as inputs for LLMs.

It is worth noting that despite the concatenation, the number of visual tokens in our method remains significantly lower than the original visual tokens generated by the vision transformer. For example, while LLaVA~\cite{liu2023visual} uses 256 visual tokens, our method utilizes only 112 visual tokens.

\begin{table*}[t]
\footnotesize
\begin{minipage}[b]{0.45\textwidth}
\centering
\setlength{\tabcolsep}{2.3pt}
{
\begin{tabular}{lcccccc}
\toprule[1.pt]
\multirow{2}{*}{\textbf{Methods}} & \textbf{LLM} & \textbf{Visual} & \multirow{2}{*}{\textbf{Conversation}} & \multirow{2}{*}{\textbf{Detail}} & \multirow{2}{*}{\textbf{Reason}} & \multirow{2}{*}{\textbf{All}} \\
& \textbf{Size} & \textbf{Tokens} \\ \midrule
 \largemodel LLaVA~\cite{liu2023visual} & \largemodel 13B & \largemodel 256 & \largemodel 83.1 & \largemodel 75.3 & \largemodel 96.5 & \largemodel 85.1 \\
 LLaVA~\cite{liu2023visual} & 7B & 256 & 70.3 & 56.6 & 83.3 & 70.1 \\ 
 LLaVA~\cite{liu2023visual}\ssymbol{2} & 7B & 256 & 78.8 & 70.2 & 91.8 & 80.4 \\ \midrule
 \rowcolor{aliceblue!60} \bf{Chat-UniVi} & 7B & \bf{112} & \bf{84.1} & \bf{74.2} & \bf{93.7} & \bf{84.2} \\
\bottomrule[1.pt]
\end{tabular}
\vspace{-.8em}
\caption{\textbf{GPT-based evaluation for image understanding.} ``\ssymbol{2}'' denotes our own re-implementation of LLaVA under our training settings (same foundation model, same image data, and same training scheme) for a fair comparison.}
\label{tab:gpt_image}
}
\end{minipage}
\hfill
\quad
\
\begin{minipage}[b]{0.52\textwidth}
\centering
\setlength{\tabcolsep}{1.6pt}
{
\begin{tabular}{lccccccc}
\toprule[1.pt]
\multirow{2}{*}{\textbf{Methods}} & {\textbf{LLM}} & \multirow{2}{*}{\textbf{Correct}} & \multirow{2}{*}{\textbf{Detail}} & \multirow{2}{*}{\textbf{Context}} & \multirow{2}{*}{\textbf{Temporal}} & \multirow{2}{*}{\textbf{Consistency}} \\
 & {\textbf{Size}} \\ \midrule
 Video-LLaMA~\cite{zhang2023video} & 7B & 39.2 & 43.6 & 43.2 & 36.4 & 35.8 \\
 LLaMA-Adapter~\cite{zhang2023llama} & 7B & 40.6 & 46.4 & 46.0 & 39.6 & 43.0 \\
 VideoChat~\cite{li2023videochat} & 7B & 44.6 & 50.0 & 50.6 & 38.8 & 44.8 \\
 Video-ChatGPT~\cite{maaz2023video} & 7B & 48.0 & 50.4 & 52.4 & 39.6 & 47.4 \\ \midrule
 \rowcolor{aliceblue!60} \bf{Chat-UniVi} & 7B & \bf{57.8} & \bf{58.2} & \bf{69.2} & \bf{47.9} & \bf{56.2} \\
\bottomrule[1.pt]
\end{tabular}
\vspace{-.6em}
\caption{\textbf{GPT-based evaluation for video understanding.} The results reported in \citet{maaz2023video} span a range from 0 to 5. To standardize the metrics, we normalize all scores to a scale of 0 to 100.}
\label{tab:gpt_video}
}
\end{minipage}
\hfill
\vspace{.2em}
\end{table*}

\begin{table*}[t]
\footnotesize
\centering
\setlength{\tabcolsep}{9.8pt}
{
\begin{tabular}{lcccccccccc}
\toprule[1.pt]
\multirow{2}{*}{\textbf{Methods}} & \multirow{2}{*}{\textbf{LLM Size}} &\multicolumn{3}{c}{\textbf{Subject}} & \multicolumn{3}{c}{\textbf{Context Modality}} & \multicolumn{2}{c}{\textbf{Grade}} &\multirow{2}{*}{\textbf{Average}} \\ 
\cmidrule(rl){3-5} \cmidrule(rl){6-8} \cmidrule(rl){9-10} & & NAT & SOC & LAN & TXT & IMG  & NO & G1-6 & G7-12 & \\ \midrule
 Random Choice~\cite{lu2022learn} & - & 40.28 & 46.13 & 29.25 & 47.45 & 40.08 & 33.66  & 39.35 & 40.67 & 39.83 \\
 Human~\cite{lu2022learn} & - & 90.23 & 84.97 & 87.48 & 89.60 & 87.50 & 88.10  & 91.59 & 82.42 & 88.40 \\ \midrule[.8pt]
 \multicolumn{4}{l}{\emph{{\textbf{Zero-shot Question Answering Accuracy (\%) }}}} \\
 \largemodel GPT-4~\cite{liu2023visual} & \largemodel 1T+ & \largemodel 84.06 & \largemodel 73.45 & \largemodel 87.36 & \largemodel 81.87 & \largemodel 70.75 & \largemodel 90.73 & \largemodel 84.69 & \largemodel 79.10 & \largemodel 82.69  \\ 
 \largemodel GPT-3~\cite{lu2022learn} & \largemodel 175B & \largemodel 75.04 & \largemodel 66.59 & \largemodel 78.00 & \largemodel 74.24 & \largemodel 65.74 & \largemodel 79.58  & \largemodel 76.36 & \largemodel 69.87 & \largemodel 74.04 \\
 LLaVA~\cite{liu2023visual}\ssymbol{2} & 7B & 47.78 & 41.96 & 53.64 & 47.90 & 44.03 & 51.92 & 49.63 & 45.29 & 48.08  \\ \midrule
 \rowcolor{aliceblue!60} \bf{Chat-UniVi} & 7B & \bf{58.61} & \bf{61.08} & \bf{61.82} & \bf{57.33} & \bf{58.25} & \bf{61.39} & \bf{62.04} & \bf{56.23} & \bf{59.96}  \\ 
 \midrule[.8pt]
 \multicolumn{4}{l}{\emph{{\textbf{Fine-tuning Question Answering Accuracy (\%) }}}} \\
 \largemodel LLaVA~\cite{liu2023visual} & \largemodel 13B & \largemodel 90.36 & \largemodel 95.95 & \largemodel 88.00 & \largemodel 89.49 & \largemodel 88.00 & \largemodel 90.66 & \largemodel 90.93 & \largemodel 90.90 & \largemodel 90.92  \\
 LLaVA~\cite{liu2023visual}\ssymbol{2} & 7B & 79.71 & 91.68 & 82.82 & 80.94 & 83.24 & 81.46 & 83.74 & 81.74 & 83.02  \\ 
 LLaMA-Adapter~\cite{zhang2023llama} & 7B & 84.37 & 88.30 & 84.36 & 83.72 & 80.32 & 86.90 & 85.83 & 84.05 & 85.19  \\  
 LLaMA-SciTune~\cite{horawalavithana2023scitune} & 7B & 84.50 & \bf{94.15} & 82.91 & 88.35 & 83.64 & \bf{88.74} & 85.05 & 85.60 & 86.11  \\ \midrule
 \rowcolor{aliceblue!60} \bf{Chat-UniVi} & 7B & \bf{88.50} & 93.03 & \bf{85.91} & \bf{88.51} & \bf{85.97} & 88.15 & \bf{88.88} & \bf{88.60} & \bf{88.78}  \\ 
\bottomrule[1.pt]
\end{tabular}
\vspace{-.5em}
\caption{\textbf{Zero-shot and fine-tuning question answering accuracy on the ScienceQA test set.} Question classes: NAT = natural science, SOC = social science, LAN = language science, TXT = text context, IMG = image context, NO = no context, G1-6 = grades 1-6, G7-12 = grades 7-12. ``\ssymbol{2}'' denotes our own re-implementation of LLaVA under our training settings for a fair comparison.}
\label{tab:scienceqa}
}
\vspace{-.8em}
\end{table*}

\subsection{Multimodal Training Scheme}
\noindent \myparagraph{Multimodal Pre-training.} 
Following the approach of previous works~\cite{liu2023visual}, our training is divided into two stages. In the first stage, we pre-train the projection matrix $\bm{W}$ while freezing both the LLM and the vision encoder. This strategic freezing of the LLM empowers our method to effectively capture semantic visual information without any discernible compromise in the performance of LLMs.

\noindent \myparagraph{Joint Instruction Tuning.} 
After completing the first stage, the model is able to understand human queries but still fails to generate reasonable and coherent linguistic responses. In the second stage, we fully fine-tune the large language model and the projection matrix $\bm{W}$ on a multimodal instruction-following dataset. This dataset is a composite of multi-turn conversations and single-turn conversations presented in a conversational format, alongside single images, multiple images, and videos as visual input. Through joint training on the mixture dataset, Chat-UniVi achieves a superior comprehension of various directives and produces more natural and dependable output. Moreover, it exhibits the distinctive ability to seamlessly process both images and videos without requiring any realignment.

\begin{table*}[t]
\footnotesize
\centering
\setlength{\tabcolsep}{9.8pt}
{
\begin{tabular}{lccccccccc}
\toprule[1.pt]
\multirow{2}{*}{\textbf{Methods}} & \multirow{2}{*}{\textbf{LLM Size}} &\multicolumn{2}{c}{\textbf{MSRVTT-QA}} & \multicolumn{2}{c}{\textbf{MSVD-QA}} & \multicolumn{2}{c}{\textbf{TGIF-QA}} & \multicolumn{2}{c}{\textbf{ActivityNet-QA}} \\ 
\cmidrule(rl){3-4} \cmidrule(rl){5-6} \cmidrule(rl){7-8} \cmidrule(rl){9-10} & & Accuracy & Score & Accuracy & Score & Accuracy & Score & Accuracy & Score \\ \midrule
 FrozenBiLM~\cite{yang2022zero} & 1B & 16.8 & - & 32.2 & - & 41.0 & -  & 24.7 & -  \\
 Video-LLaMA~\cite{zhang2023video} & 7B & 29.6 & 1.8 & 51.6 & 2.5 & - & - & 12.4 & 1.1 \\
 LLaMA-Adapter~\cite{zhang2023llama} & 7B & 43.8 & 2.7 & 54.9 & 3.1 & - & - & 34.2 & 2.7 \\
 VideoChat~\cite{li2023videochat} & 7B & 45.0 & 2.5 & 56.3 & 2.8 & 34.4 & 2.3 & 26.5 & 2.2 \\
 Video-ChatGPT~\cite{maaz2023video} & 7B & 49.3 & 2.8 & 64.9 & 3.3 & 51.4 & 3.0 & 35.2 & 2.7 \\ \midrule
 \rowcolor{aliceblue!60} \bf{Chat-UniVi} & 7B & \bf{55.0} & \bf{3.1} & \bf{69.3} & \bf{3.7} & \bf{69.0} & \bf{3.8} & \bf{46.1} & \bf{3.3} \\
\bottomrule[1.pt]
\end{tabular}
\vspace{-.6em}
\caption{\textbf{Zero-shot video question answering accuracy.} We follow the evaluation protocol in \citet{maaz2023video}, \ie, employing GPT-assisted evaluation to assess the capabilities of models. ``Score'' denotes the confidence score from 0 to 5 assigned by the GPT model.}
\label{tab:videoqa}
}
\end{table*}

\begin{table*}[t]
\footnotesize
\centering
\setlength{\tabcolsep}{6pt}
{
\begin{tabular}{lcccccccccc}
\toprule[1.pt]
\multirow{2}{*}{\textbf{Methods}} & \multirow{2}{*}{\textbf{LLM Size}} & \multicolumn{3}{c}{\textbf{Random (POPE-R)}} & \multicolumn{3}{c}{\textbf{Popular (POPE-P)}} & \multicolumn{3}{c}{\textbf{Adversarial (POPE-A)}} \\ 
\cmidrule(rl){3-5} \cmidrule(rl){6-8} \cmidrule(rl){9-11} & & Accuracy & F1-Score & Yes & Accuracy  & F1-Score & Yes & Accuracy & F1-Score  & Yes \\ \midrule
 \largemodel LLaVA~\cite{liu2023visual} & \largemodel 13B & \largemodel 64.12 & \largemodel 73.38 & \largemodel 83.26 & \largemodel 63.90 & \largemodel 72.63 & \largemodel 81.93 & \largemodel 58.91 & \largemodel 69.95 & \largemodel 86.76 \\
 \largemodel MiniGPT-4~\cite{zhu2023minigpt} & \largemodel 13B & \largemodel 79.67 & \largemodel 80.17 & \largemodel 52.53 & \largemodel 69.73 & \largemodel 73.02 & \largemodel 62.20 & \largemodel 65.17 & \largemodel 70.42 & \largemodel 67.77  \\ 
 \largemodel InstructBLIP~\cite{dai2023instructblip} & \largemodel 13B & \largemodel 88.57 & \largemodel 89.27 & \largemodel 56.57 & \largemodel 82.77 & \largemodel 84.66 & \largemodel 62.37 & \largemodel 72.10 & \largemodel 77.32 & \largemodel 73.03  \\ 
 MultiModal-GPT~\cite{gong2023multimodal} & 7B & 50.10 & 66.71 & 99.90 & 50.00 & 66.67 & 100.00  & 50.00 & 66.67 & 100.00 \\
 mPLUG-Owl~\cite{ye2023mplug} & 7B & 53.97 & 68.39 & 95.63 & 50.90 & 66.94 & 98.57 & 50.67 & 66.82 & 98.67 \\
 LLaVA~\cite{liu2023visual}\ssymbol{2} & 7B & 72.16 & 78.22 & 76.29 & 61.37 & 71.52 & 85.63 & 58.67 & 70.12 & 88.33  \\ 
 \midrule
 \rowcolor{aliceblue!60} Chat-UniVi w/o multi-scale & 7B & 73.88 & 79.30 & 74.63 &  56.36 & 69.01 & 90.83 & 55.63 & 68.67 & 91.63  \\ 
 \rowcolor{aliceblue!60} \bf{Chat-UniVi w/ multi-scale} & 7B & \bf{85.19} & \bf{86.05} & \bf{54.67} & \bf{69.50} & \bf{74.39} & \bf{69.10} & \bf{64.97} & \bf{71.54} & \bf{73.10}  \\ 
\bottomrule[1.pt]
\end{tabular}
\vspace{-.6em}
\caption{\textbf{Zero-shot object hallucination evaluation on the COCO validation set.}  We report the results of the polling-based object probing evaluation (POPE). ``Yes'' represents the proportion of positive answers that the model outputs. ``\ssymbol{2}'' denotes our own re-implementation of LLaVA under our training settings (same foundation model, same image data, and same training scheme) for a fair comparison.}
\label{tab:pope}
}
\vspace{-.8em}
\end{table*}

\section{Experiments}
\subsection{Experimental Setup}\label{Experimental Setup0}
\noindent  \myparagraph{Model Settings.} We adopt the vision encoder of CLIP~(ViT-L/14)~\cite{radford2021learning} as the visual foundation model. Besides, we chose the Vicuna-v1.5 model~\cite{vicuna}, which consists of 7B parameters, as our language foundation model.

\noindent \myparagraph{Data and Training Details.}
For the multimodal pre-training stage, we utilize the image-caption pairs from various datasets, including COCO~\cite{chen2015microsoft} and CC3M-595K screened from CC3M~\cite{sharma2018conceptual} by LLaVA~\cite{liu2023visual}. We pre-train Chat-UniVi for one epoch with a batch size of 128, employing the AdamW~\cite{kingma2014adam,loshchilov2017decoupled} optimizer with a cosine schedule. The learning rate is set to 2e-3, and the warm-up rate is 0.03. For the joint instruction tuning stage, we incorporate multimodal instruction data from multiple sources: (i) multimodal in-context instruction datasets, such as MIMIC-IT~\cite{li2023otter,antol2015vqa,hudson2019gqa}, (ii) visual instruction datasets, such as LLaVA, (iii) video instruction data from Video-ChatGPT~\cite{maaz2023video}. All input images or frames are resized to $224\times224$. We train Chat-UniVi for 2 epochs with a batch size of 128, and the learning rate is set to 2e-5.

\begin{table*}[t]
\footnotesize
\setlength{\tabcolsep}{10.3pt}
{
\centering
\begin{tabular}{lcccccccccc}
\toprule[1.pt]
\multirow{2}{*}{\textbf{Methods}} & \multicolumn{4}{c}{\textbf{Image Understanding}}  & \multicolumn{5}{c}{\textbf{Video Understanding}} \\
\cmidrule(rl){2-5} \cmidrule(rl){6-10}
& {Conversation} & {Detail} & {Reason} & {All} & {Correct} & {Detail} & {Context} & {Temporal} & {{Consistency}} 
 \\ \midrule
 Only Image & 84.0 & 69.3 & 89.3 & 81.5 & 43.4 & 48.6 & 56.8 & 36.6 & 46.2 \\
 Only Video & 72.7 & 55.8 & 71.5 & 66.8 & 57.4 & \bf{58.8} & 69.0 & 47.0 & 56.0 \\
 Image + Video & 45.5 & 31.3 & 76.1 & 50.9 & 51.2 & 55.6 & 64.8 & 40.3 & 50.4 \\
 Video + Image & 79.0 & 69.2 & 88.5 & 79.1 & 45.6 & 49.8 & 58.2 & 38.8 & 47.8 \\ \midrule
 \rowcolor{aliceblue!60} \bf{Image \& Video} & \bf{84.1} & \bf{74.2} & \bf{93.7} & \bf{84.2} & \bf{57.8} & 58.2 & \bf{69.2} & \bf{47.9} & \bf{56.2} \\
\bottomrule[1.pt]
\end{tabular}
\vspace{-.6em}
\caption{\textbf{Ablation study about instruction tuning scheme.} ``Only Image'' indicates training solely on image data. ``Image + Video'' means training on image data followed by fine-tuning on video data. ``Image \& Video'' denotes training on a mixed dataset.}
\label{tab:ab_tuning}
}
\vspace{1.3em}

\footnotesize
\begin{minipage}[b]{0.48\textwidth}
\centering
\setlength{\tabcolsep}{3.2pt}
{
\begin{tabular}{ccc|c|cccc}
\toprule[1.pt]
\bm{$C_1$} &\bm{$C_2$} & \bm{$C_3$} & \textbf{Visual Tokens} &{\textbf{Conversation}} & {\textbf{Detail}} & {\textbf{Reason}} & {\textbf{All}} \\ \midrule
 16 & 8 & 4 & 28 & 78.6 & 69.0 & \bf{95.1} & 81.1 \\  
 32 & 16 & 8 & 56 & 82.7 & 67.2 & 94.5 & 81.6 \\  
 \rowcolor{aliceblue!60} 64 & 32 & 16 & 112 & \bf{84.1} & \bf{74.2} & 93.7 & \bf{84.2} \\ 
 128 & 64 & 32 & 224 & 79.8 & 68.7 & 83.8 & 79.8 \\ 
\bottomrule[1.pt]
\end{tabular}
\vspace{-.6em}
\caption{\textbf{Ablation study about the number of spatial visual clusters.} ``$C_1$'', ``$C_2$'', and ``$C_3$'' denote the number of clusters at the first step, the second step, and the last step, respectively.}
\label{tab:ab_clu}
}
\end{minipage}
\hfill
\quad
\
\begin{minipage}[b]{0.5\textwidth}
\centering
\setlength{\tabcolsep}{3.2pt}
{
\begin{tabular}{ccccccc}
\toprule[1.pt]
 {\textbf{Clustering Ratio}} & {\textbf{Correct}} & {\textbf{Detail}} &{\textbf{Context}} & {\textbf{Temporal}} & {\textbf{Consistency}} \\ \midrule
 $1/M$ & 51.2 & 41.8 & 47.6 & 28.0 & 42.2 \\ 
 $1/32$ & 57.2 & 58.0 & \bf{69.6} & 45.8 & 54.2 \\
 \rowcolor{aliceblue!60} $1/16$ & \bf{57.8} & \bf{58.2} & 69.2 & \bf{47.9} & 56.2 \\ 
 $1/8$ & 56.8 & \bf{58.2} & 68.0 & 46.2 & \bf{57.8} \\
\bottomrule[1.pt]
\end{tabular}
\vspace{-.6em}
\caption{\textbf{Ablation study about the number of temporal visual clusters.} ``$M$'' is the frame length. ``$1/M$'' denotes that the model directly consolidates all frames into a single event.}
\label{tab:ab_event}
}
\end{minipage}
\hfill
\vspace{-1.2em}
\end{table*}

\begin{figure*}[ht]
\centering
\includegraphics[width=1.\textwidth]{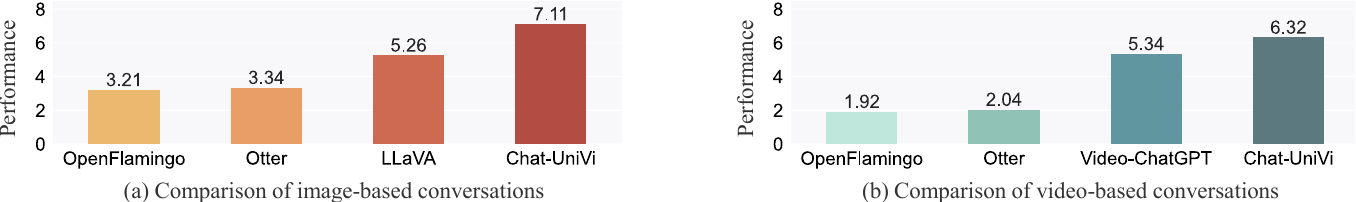}
\vspace{-1.5em}
\caption{\textbf{Human evaluations.} In 30 image conversation scenarios and 30 video conversation scenarios, the evaluators rate the model on a scale of 0 to 10 based on its multimodal conversation performance. Finally, we use the average score as the final model score.}
\label{fig:user_study}
\vspace{-.6em}
\end{figure*}

\subsection{GPT-based evaluation}
\noindent \myparagraph{Image Understanding.} To quantitatively measure the image understanding capability, we report the GPT-4 evaluation results in \cref{tab:gpt_image}. Following \citet{liu2023visual,zhang2023llavar}, we employ 90 questions based on 30 COCO validation images, covering various aspects, including conversation, detail description (Detail), and complex reasoning (Reason). We utilize the GPT-4 model to evaluate the outputs of the model in these three aspects, as well as provide an overall score. For a comprehensive description of image understanding metrics, please refer to the appendix. As shown in \cref{tab:gpt_image}, Chat-UniVi uses fewer visual tokens while achieving superior performance. Notably, our method, even as a 7B model, can achieve the performance level of a 13B model, demonstrating the effectiveness of our method.

\noindent \myparagraph{Video Understanding.} To quantitatively measure the video understanding capability, we report the GPT evaluation results in \cref{tab:gpt_video}. Following \citet{maaz2023video}, we employ a test set based on the ActivityNet dataset~\cite{caba2015activitynet} and utilize the GPT-3.5 model to assign a relative score to the outputs of the model in the following five aspects: Correctness of Information (Correct), Detail Orientation (Detail), Contextual Understanding (Context), Temporal Understanding (Temporal), and Consistency. Please refer to the appendix for more details. As shown in \cref{tab:gpt_video}, Chat-UniVi, even as a unified model, significantly surpasses recently proposed state-of-the-art methods that exclusively focus on video, which demonstrates the effectiveness of our method.

\begin{figure*}[t]
\centering
\includegraphics[width=1\textwidth]{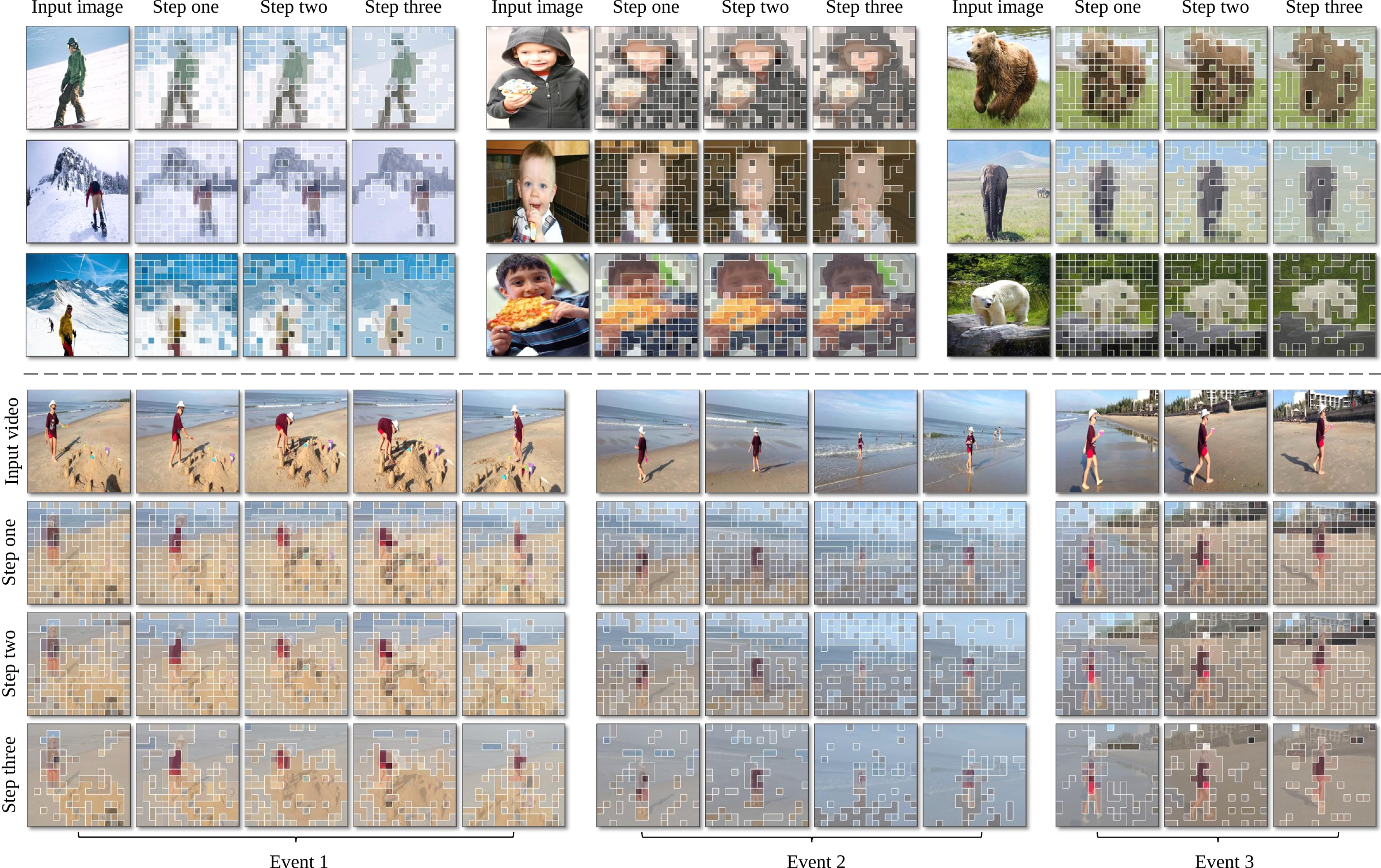}
\vspace{-1.3em}
\caption{\textbf{Visualization of the dynamic visual tokens.} For clarity in observation, we map the dynamic visual tokens of the video back to each frame for visualization. Please refer to the appendix for additional visualizations and conversation examples of our model.}
\vspace{-.6em}
\label{fig:visualization}
\end{figure*}

\subsection{Question-Answer Evaluation}
\myparagraph{ScienceQA Performance.} ScienceQA~\cite{lu2022learn} is a multimodal science question-answering dataset comprising 21k multiple-choice questions. Each example in ScienceQA contains a visual context, a textual context, a question, and multiple options. We report both zero-shot and fine-tuning results in \cref{tab:scienceqa}. As shown in \cref{tab:scienceqa}, Chat-UniVi shows competitive performance across all metrics. Notably, Chat-UniVi outperforms LLaMA-SciTune~\cite{horawalavithana2023scitune}, a model specifically tailored for science question answering, which fully demonstrates the superiority of our method.

\noindent \myparagraph{Zero-shot Video-question Answering Performance.} 
In \cref{tab:videoqa}, we show the zero-shot video-question answering performance on several commonly used open-ended question-answer datasets, including MSRVTT-QA~\cite{xu2017video}, MSVD-QA~\cite{xu2017video}, TGIF-QA FrameQA~\cite{jang2017tgif}, and ActivityNet-QA~\cite{yu2019activitynet}. Our evaluation protocol follows that of \citet{maaz2023video}, utilizing GPT-assisted evaluation to assess the capabilities of models. As shown in \cref{tab:videoqa}, Chat-UniVi outperforms the recently proposed state-of-the-art methods, \eg, FrozenBiLM~\cite{yang2022zero}, across various datasets.

\subsection{Object Hallucination Evaluation}
In \cref{tab:pope}, we report the results of the polling-based object probing evaluation~\cite{li2023evaluating} (POPE). For details of the polling-based object probing evaluation, please refer to the appendix. As shown in \cref{tab:pope}, Chat-UniVi outperforms the recently proposed state-of-the-art methods. Moreover, we find that multi-scale representation improves the ability to resist hallucinations. It is worth noting that, as a 7B model, our method even outperforms the 13B model, such as MiniGPT-4. We attribute this success to the multi-scale representation that equips our method to perceive both high-level semantic concepts and low-level visual appearance. 

\subsection{Ablative Analysis}
\noindent \myparagraph{Effect of the Tuning Scheme.} 
In \cref{tab:ab_tuning}, we provide the ablation study on the instruction tuning scheme. We find that visual instruction tuning using only one type of medium, such as images, results in a decrease in comprehension of another medium, such as videos. However, pre-training on one medium and fine-tuning on another leads to knowledge degradation from the pre-training stage. In contrast, our joint training strategy, which involves training on a mixed dataset of images and videos, endows the model with the capability to process both types of visual inputs. Among all tuning schemes, joint training consistently achieves the highest performance, confirming its effectiveness.

\noindent \myparagraph{Effect of the Number of Spatial Visual Clusters.} 
To explore the influence of the number of spatial visual clusters, we provide the ablation results in \cref{tab:ab_clu}. We find that a smaller number of visual clusters may decrease the capacity to grasp fine visual details, whereas a larger number of visual clusters may introduce redundancy and potentially reduce the overall performance of the model. To strike a balance between detailed understanding and model learning complexity, we set the number of clusters at the three levels to 64, 32, and 16 respectively in practice.

\noindent \myparagraph{Effect of the Number of Temporal Visual Clusters.} Videos vary in length, with longer videos typically containing more events. Therefore, in Chat-UniVi, the number of temporal visual clusters is determined proportionally based on the number of input video frames. As shown in \cref{tab:ab_event}, we find that a smaller clustering ratio may result in the loss of crucial temporal information within the video. Conversely, a larger clustering ratio increases the computational overhead of the model. We observe that the model performs optimally when the clustering ratio is set to $1/16$. Therefore, in practice, we adopt a default temporal clustering ratio of $1/16$ for optimal performance.

\subsection{Qualitative Analysis}
\noindent \myparagraph{Human Evaluation.} In our evaluation, we manually assess the performance of Chat-UniVi and baselines in 30 image conversation scenarios and 30 video conversation scenarios. The results are presented in \cref{fig:user_study}. OpenFlamingo~\cite{awadalla2023openflamingo}, derived from Flamingo~\cite{alayrac2022flamingo}, and Otter~\cite{li2023otter}, an in-context instruction tuning variant of OpenFlamingo, are also included in our comparison. As shown in \cref{fig:user_study}, we find that methods based on Flamingo exhibit limitations in their ability to comprehend videos. This limitation is attributed to their use of a query transformer to extract a fixed number of visual tokens from videos of varying lengths, which hinders their effectiveness in modeling temporal comprehension. In contrast, Chat-UniVi, functioning as a unified model, not only outperforms methods built upon the Flamingo but also surpasses models specifically designed for image and video.

\noindent \myparagraph{Visualization of the Dynamic Visual Tokens.} We provide the visualization in \cref{fig:visualization} and invite readers to explore more visualizations in the appendix. It is important to emphasize that our proposed token merging method operates without the need for object outline labels. As shown in \cref{fig:visualization}, the proposed dynamic visual tokens effectively generalize objects and backgrounds. This capability enables Chat-UniVi to reconcile the intricate spatial nuances of images with the broader temporal understanding required for videos with a limited number of visual tokens.

\section{Conclusion}
In this paper, we introduce Chat-UniVi, a unified multimodal large language model designed to comprehend and engage in conversations about both images and videos. To seamlessly bridge the intricate spatial nuances of images with the broader temporal understanding required for videos, we propose a unified representation framework employing dynamic visual tokens. This representation leverages DPC-KNN to progressively cluster visual tokens and provides multi-scale features. More encouragingly, Chat-UniVi is trained on a mixed dataset encompassing both images and videos, enabling it to be directly applicable to tasks involving both media types without requiring any modifications. Extensive experimental results demonstrate that Chat-UniVi, as a unified model, consistently surpasses even methods exclusively designed for images or videos.

\noindent \myparagraph{Acknowledgements.} This work was supported by the National Key R\&D Program of China (2022ZD0118101), Nature Science Foundation of China (No.62202014), and
Shenzhen Basic Research Program (No.JCYJ20220813151736001).

{
    \small
    \bibliographystyle{ieeenat_fullname}
    \bibliography{main}
}

\clearpage
\appendix

\renewcommand{\thefootnote}{\fnsymbol{footnote}}

\renewcommand{\thetable}{\Alph{table}}
\renewcommand{\theequation}{\Alph{equation}}
\renewcommand{\thefigure}{\Alph{figure}}

\setcounter{table}{0}
\setcounter{section}{0}
\setcounter{figure}{0}
\setcounter{equation}{0}

\clearpage
\myparagraph{Abstract} This appendix provides additional discussions~(Appendix~\ref{appendix:Additional Discussions}), implementation details~(Appendix~\ref{Experimental Setup1}), several additional experiments~(Appendix~\ref{appendix:Additional Experiments}), additional visualization results~(Appendix~\ref{appendix:Additional Visualization Results}), more qualitative analysis~(Appendix~\ref{appendix:Additional Qualitative Analysis}), and details of quantitative evaluations~(Appendix~\ref{appendix:Details of Quantitative Evaluations}).

\section{Additional Discussions}\label{appendix:Additional Discussions}
\subsection{Comparison of Chat-UniVi and Other Multimodal Methods}
Existing methods often focus exclusively on either image or video inputs. Recently, there have also been some methods~\cite{alayrac2022flamingo,wu2023next,chen2023x} that support both images and videos, and they can be broadly divided into two classes. 

\begin{table*}[thb]
\centering
{
\begin{tabular}{ccccc}
\toprule[.9pt]
\multirow{2}{*}{\textbf{Type}} & \multirow{2}{*}{\textbf{Methods}} & \textbf{Variable} & \textbf{Unified} & \textbf{Benefit from} \\
&  & \textbf{Length Features} & \textbf{Visual Encoder} & \textbf{Joint Training}
 \\ \midrule
 \multirow{2}{*}{Q-former based methods} & Flamingo & \multirow{2}{*}{\ding{56}} & \multirow{2}{*}{\ding{52}} & \multirow{2}{*}{\textbf{--}} \\ 
  & OpenFlamingo, Otter\\ \midrule
 Multi-encoder methods  & X-LLM, NExT-GPT & \textbf{--} & \ding{56} & \ding{56} \\ \midrule
 \rowcolor{aliceblue!60} Unified methods & Chat-UniVi & \ding{52} & \ding{52} & \ding{52} \\
\bottomrule[.9pt]
\end{tabular}
}
\vspace{-.4em}
\caption{\textbf{Comparison with other methods.} ``\ding{56}'' denotes that the model does not have this property. ``\ding{52}'' denotes that the model has this property. ``\textbf{--}'' indicates a temporary lack of experimental evidence.}
\label{tab:apendix_methods}
\end{table*}

\begin{table*}[th]
\centering
{
\begin{tabular}{lcccccc}
\toprule[.9pt]
\multirow{2}{*}{\textbf{Methods}} & \multirow{2}{*}{\textbf{Parameter-free}} & \multirow{2}{*}{\textbf{Video Input}}  & \multicolumn{4}{c}{\textbf{Image Understanding}}   \\
\cmidrule(rl){4-7} 
& & & {Conversation} & {Detail} & {Reason} & {All}
 \\ \midrule
 \citet{ma2023image} & \ding{56} & \ding{56} & 71.8 & 60.9 & 91.6 & 75.0  \\ \midrule
 \rowcolor{aliceblue!60} Chat-UniVi & \ding{52} & \ding{52} & \bf{84.1} & \bf{74.2} & \bf{93.7} & \bf{84.2}  \\
\bottomrule[.9pt]
\end{tabular}
}
\vspace{-.4em}
\caption{\textbf{Comparison of Chat-UniVi and another token clustering method.} ``\ding{56}'' denotes that the model does not have this property. ``\ding{52}'' denotes that the model has this property.}
\label{tab:appendix_cluster}
\end{table*}

\begin{itemize}
\item \textbf{Q-former based methods.} The first class of methods uses a query transformer to extract a fixed number of tokens for each image and video. These methods are exemplified by Flamingo~\cite{alayrac2022flamingo}, OpenFlamingo~\cite{awadalla2023openflamingo}, and Otter~\cite{li2023otter}. However, videos vary in length, posing a challenge for these methods, as they extract a fixed number of visual tokens from each video, limiting their ability to effectively capture temporal comprehension. Human evaluation results (see Fig.~\ref{fig:user_study}) also substantiate that these methods struggle to strike a balance between image and video comprehension.

\item \textbf{Multi-encoder methods.} The second category of methods employs separate pre-trained image and video encoders to process images and videos independently. Prominent examples of this approach include X-LLM~\cite{chen2023x} and NExT-GPT~\cite{wu2023next}. However, these methods introduce redundancy within the model and present difficulties when trained jointly. Most importantly, this approach does not leverage the advantages of joint training with both image and video data. Consequently, they do not align with our primary objective of developing a unified vision-language model.
\end{itemize}

In contrast to the previous works, Chat-UniVi uniformly represents images and videos using multi-scale dynamic visual tokens. The proposed Chat-UniVi has two compelling advantages:

\begin{itemize}
\item \textbf{Variable length video features.} In Chat-UniVi, the number of temporal visual clusters is determined proportionally based on the number of input video frames. In contrast to the Q-former based methods, Chat-UniVi allocates a greater number of visual tokens to longer videos. Therefore, our method is better suited for variable-length video understanding. 

\item \textbf{Unified visual encoder.} Chat-UniVi employs a shared visual encoder to consistently process both images and videos. In contrast to multi-encoder methods, our method eliminates the need for introducing redundant parameters and streamlines the training process.

\item \textbf{Benefit from joint training.} Due to the unified representation framework for both images and videos, Chat-UniVi can be trained on mixed datasets that include both images and videos. This allows for direct application to tasks involving both images and videos. Most importantly, we find that this joint training strategy can simultaneously enhance the model's understanding of both images and videos. Experimental results are shown in Tab.~\ref{tab:ab_tuning}.
\end{itemize}

In Tab.~\ref{tab:apendix_methods}, we show the comparison of Chat-UniVi and other methods. For Q-former based methods, the advantages of joint training are not shown, and even the performance of the model may affect each other when multiple datasets are mixed~\cite{alayrac2022flamingo}. However, the potential to benefit from joint training cannot be ruled out. In addition, the multi-encoder method can also select a video encoder that can encode dynamic length features.

\subsection{Comparison of Chat-UniVi and Other Clustering Transformer Methods}
There have also been recent methods~\cite{ma2023image,xu2022groupvit,zeng2022not,jin2023video} to explore the role of token clustering within the transformer framework. However, none of these methods can be directly extended to video, and additional parameters need to be trained. We summarize the advantages of our method as follows:

\begin{itemize}
\item \textbf{Supporting video input.} In contrast to other methods, Chat-UniVi extends the tokens clustering method to incorporate video inputs, achieving the integration of image and video representations for the first time. Our work is the first to demonstrate that this unified representation can reconcile the intricate spatial details of images with the broader temporal understanding required for videos.

\item \textbf{Without parameters.} Our clustering method is parameter-free and therefore requires no training. Interestingly, we find that this parameter-free clustering method serves as the linchpin to the success of our model. As shown in Tab.~\ref{tab:appendix_cluster},  the performance of the clustering method with training parameters is significantly inferior to the parameter-free clustering method we propose. We attribute this phenomenon to the gradient instability in multimodal conversation training, which hinders the convergence of parameterized methods.
\end{itemize}

\begin{table*}[h]
\centering
{
{
{
\begin{tabular}{lcccccccc}
\toprule[1.pt]
 \multirow{2}{*}{{Methods}} & \multicolumn{2}{c}{{Time Complexity}} 
  & \multicolumn{3}{c}{{Image Inference}}  & \multicolumn{3}{c}{{Video Inference}} \\ \cmidrule(rl){2-3} \cmidrule(rl){4-6} \cmidrule(rl){7-9}
  & Spatial & Temporal & Merging (s) & All (s) & Memory (M) & Merging (s) & All (s) & Memory (M) \\ \midrule
 LLaVA & - & - & {0} & {2.3116} & {15673} & {\ding{56}} & {\ding{56}} & {\ding{56}} \\
 \rowcolor{aliceblue!60} Ours & $\mathcal{O}(L^2D)$ & $\mathcal{O}(M^2D)$ & {0.0027} & {2.2722} & {15443} & {0.0174} & {4.4040} & {16533} \\
\bottomrule[1.pt]
\end{tabular}
}
}}
\vspace{-.4em}
\caption{{\textbf{Runtime and memory complexity analysis.} $L$, $D$, and $M$ denote the number of vanilla visual tokens, the feature dimension, the frame length, respectively. ``\ding{56}'' denotes that the method does not have this property.}}
\label{tab:efficiency}
\end{table*}

\begin{table*}[t]
\centering
{
\begin{tabular}{lcccc}
\toprule[.9pt]
\multirow{2}{*}{\textbf{Datasets}} & \multirow{2}{*}{\textbf{Image Inputs}} & \multirow{2}{*}{\textbf{Video Inputs}} & \textbf{Multi-turn} & \textbf{Number of} \\ 
&  &  & \textbf{Conversations} & \textbf{Conversations} \\ \midrule
 \multicolumn{4}{l}{\emph{\color{gray}{\textbf{Multimodal Pre-training Stage}}}} \\ 
 CC3M-595K & \ding{52} & \ding{56} & \ding{56} & 595K \\
 COCO & \ding{52} & \ding{56} & \ding{56} & 956K \\ \midrule
 \multicolumn{4}{l}{\emph{\color{gray}{\textbf{Joint Instruction Tuning Stage}}}} \\
 LLaVA-instruct-150K & \ding{52} & \ding{56} & \ding{52} & 150K \\
 MIMIC-IT-399K\ssymbol{3} & \ding{52} & \ding{56} & \ding{56} & 399K \\
 Video-ChatGPT-instruct & \ding{56} & \ding{52} & \ding{56} & 100K \\
 
\bottomrule[.9pt]
\end{tabular}
\vspace{-.4em}
\caption{\textbf{Description of training data.} ``\ding{56}'' denotes that the dataset does not have this property. ``\ding{52}'' denotes that the dataset has this property. ``\ssymbol{3}'' represents the dataset filtered from MIMIC-IT, containing exclusively image data. In order to further filter the training data, we also delete the duplicate data in LLaVA-instruct-150K and MIMIC-IT.}
\label{tab:appendix_data}
}
\end{table*}

\begin{table*}[t]
\centering
{
\begin{tabular}{lcccccccccc}
\toprule[.9pt]
\multirow{2}{*}{\textbf{Methods}} & \multicolumn{4}{c}{\textbf{Image Understanding}}  & \multicolumn{5}{c}{\textbf{Video Understanding}} \\
\cmidrule(rl){2-5} \cmidrule(rl){6-10}
& {Conversation} & {Detail} & {Reason} & {All} & {Correct} & {Detail} & {Context} & {Temporal} & {{Consistency}} .
 \\ \midrule
 LoRA & 76.1 & 68.6 & 82.4 & 75.8 & 52.8 & 55.0 & 63.8 & 42.6 & 53.8 \\ \midrule
 \rowcolor{aliceblue!60} Full fine-tuning & \bf{84.1} & \bf{74.2} & \bf{93.7} & \bf{84.2} & \bf{57.8} & \bf{58.2} & \bf{69.2} & \bf{47.9} & \bf{56.2} \\
\bottomrule[.9pt]
\end{tabular}
}
\vspace{-.4em}
\caption{\textbf{Comparison between the LoRA and full fine-tuning.} ``Detail'' denotes the ``Detail Description'' in the context of image understanding or ``Detail Orientation'' in the context of video understanding. For image understanding, ``Reason'' denotes the ``Complex Reasoning''. For video understanding, ``Correct'', ``Context'', and ``Temporal'' stand for ``Correctness of Information'', ``Contextual Understanding'', and ``Temporal Understanding'', respectively.}
\label{tab:appendix_lora}
\vspace{1.2em}

{
\begin{tabular}{lcccccccccc}
\toprule[.9pt]
\multirow{2}{*}{\textbf{Methods}} & \multicolumn{4}{c}{\textbf{Image Understanding}}  & \multicolumn{5}{c}{\textbf{Video Understanding}} \\
\cmidrule(rl){2-5} \cmidrule(rl){6-10}
& {Conversation} & {Detail} & {Reason} & {All} & {Correct} & {Detail} & {Context} & {Temporal} & {{Consistency}} 
 \\ \midrule
 EVA-CLIP & 80.0 & 74.7 & 91.2 & 82.1 & 57.2 & 58.8 & 67.8 & 45.7 & 54.6 \\ \midrule
 \rowcolor{aliceblue!60} Openai-CLIP & \bf{84.1} & \bf{74.2} & \bf{93.7} & \bf{84.2} & \bf{57.8} & \bf{58.2} & \bf{69.2} & \bf{47.9} & \bf{56.2} \\
\bottomrule[.9pt]
\end{tabular}
\vspace{-.4em}
\caption{\textbf{Comparison between the EVA CLIP and the Openai CLIP.} We choose EVA-CLIP (ViT-G), which has a similar number of parameters as Openai-CLIP (ViT-L/14), for the experiment.}
\label{tab:appendix_eva}
}
\end{table*}

\subsection{Runtime and Memory Complexity}
As shown in Tab.~\ref{tab:efficiency}, the time and memory costs of our clustering algorithm are negligible compared to those of the large language model.

\subsection{Limitations and Future Work}
In this section, we delineate the limitations of our work and outline avenues for future research.

\noindent \myparagraph{The Enduring Impact of Large Language Models.}
Our method leverages the strength of pre-trained Large Language Models, and as a consequence, also inherits their vulnerabilities.

\begin{itemize}
    \item \textbf{Hallucination.} While our experiments (see Tab.~\ref{tab:pope}) demonstrate the effectiveness of our method in addressing hallucinations, it is important to acknowledge that the issue of hallucinations in LLMs remains a challenge yet to be fully resolved. The phenomenon of illusory responses in LLMs can result in unsupported conjectures during open multimodal conversations, and addressing this issue has the potential to significantly expedite advancements in the field. For a more in-depth exploration of common weaknesses observed in large LLMs, please refer to \citet{brown2020language,rae2021scaling}.
    
    \item \textbf{Long sequence processing.} Transformer-based language models often exhibit suboptimal generalization when confronted with test sequences considerably longer than their training data~\cite{press2021train}. This becomes particularly evident in multi-turn conversations, where the model may exhibit forgetfulness of prior conversational context, resulting in erroneous responses. Simultaneously, we find a decline in model performance when multiple videos are inputted, which could also be attributed to constraints associated with sequence length.
    
    \item \textbf{Prompt sensitivity.} In-context learning has demonstrated disconcerting sensitivity to various aspects of demonstrations, including prompt formats~\cite{zhao2021calibrate}. Notably, different prompt formats can yield entirely contradictory output results. Finding a solution to this issue holds the potential to greatly accelerate progress in the field.
    
\end{itemize}

\noindent \myparagraph{Natural Language Output.}
Natural language serves as a robust and adaptable input/output interface for describing visual tasks to the model, facilitating the generation of outputs, or estimating conditional probabilities for potential outcomes. However, it may prove to be a less convenient interface for tasks that require conditioning on or predicting more structured outputs, such as bounding boxes, as well as for generating dense pixel predictions. Besides, the flexibility of the natural language output also makes it difficult to evaluate the performance of the model.

\noindent \myparagraph{More Modalities.}
Future work can explore alternative modalities, such as audio, in addition to visual inputs. The incorporation of multiple modalities holds the promise of broadening the spectrum of tasks that the model can address, and it has the potential to enhance their performance by leveraging synergies among these various modalities. For example, contemplating audio information alongside video processing can significantly augment the video understanding of the model.

\section{Implementation Details}\label{Experimental Setup1}
\noindent \myparagraph{Data Details.} For the multimodal pre-training stage, we utilize the image-caption pairs from various datasets, including COCO~\cite{chen2015microsoft} and CC3M-595K screened from CC3M~\cite{sharma2018conceptual} by LLaVA~\cite{liu2023visual}. All input images are resized to $224\times224$. For the joint instruction tuning stage, we incorporate multimodal instruction data from multiple sources: (i) multimodal in-context instruction datasets, such as MIMIC-IT~\cite{li2023otter,antol2015vqa,hudson2019gqa}, (ii) visual instruction datasets, such as LLaVA, (iii) video instruction data from Video-ChatGPT~\cite{maaz2023video}. In order to further filter the training data, we delete the duplicate data in LLaVA-instruct-150K and MIMIC-IT, and delete the video data in MIMIC-IT. This dataset is a composite of multi-turn conversations and single-turn conversations presented in a conversational format, alongside single images, multiple images, and videos as visual input. For each video, we select 64 frames as input for the model. All input images or frames are resized to $224\times224$. We provide a detailed description of the training data in Tab.~\ref{tab:appendix_data}.

\noindent \myparagraph{Model Settings.} Following previous works~\cite{liu2023visual}, we adopt the vision encoder of CLIP~(ViT-L/14)~\cite{radford2021learning} as the visual foundation model. We chose an instruction-tuned variant of LLaMA2~\cite{touvron2023llama2}, \ie, Vicuna~\cite{vicuna}, as our language foundation model. Specifically, we utilize the Vicuna-v1.5 model, comprised of 7B parameters.

\noindent \myparagraph{Training Hyperparameters.}
For the multimodal pre-training stage, we pre-train Chat-UniVi for one epoch with a batch size of 128, employing the AdamW optimizer with a cosine schedule. The learning rate is set to 2e-3, and the warm-up rate is 0.03. For the joint instruction tuning stage, we train Chat-UniVi for 2 epochs with a batch size of 128, and the learning rate is set to 2e-5, employing the AdamW optimizer with a cosine schedule. The warm-up rate is set to 0.03.

\noindent \myparagraph{ScienceQA Fine-tuning Settings.} We start with a pre-trained model to fine-tune. We fine-tune the model for 9 epochs with a batch size of 32, employing the AdamW optimizer with a cosine schedule. The learning rate is set to 2e-5, and the warm-up rate is 0.03. 

\begin{table*}[t]
\centering
{
\begin{tabular}{lcccccccccc}
\toprule[.9pt]
\multirow{2}{*}{\textbf{Methods}} & \multicolumn{4}{c}{\textbf{Image Understanding}}  & \multicolumn{5}{c}{\textbf{Video Understanding}} \\
\cmidrule(rl){2-5} \cmidrule(rl){6-10}
& {Conversation} & {Detail} & {Reason} & {All} & {Correct} & {Detail} & {Context} & {Temporal} & {{Consistency}} 
 \\ \midrule
 Single-scale & 70.5 & 63.4 & 88.3 & 74.2 & 54.6 & 56.4 & 65.8 & 42.1 & 52.2 \\ \midrule
 \rowcolor{aliceblue!60} Multi-scale & \bf{84.1} & \bf{74.2} & \bf{93.7} & \bf{84.2} & \bf{57.8} & \bf{58.2} & \bf{69.2} & \bf{47.9} & \bf{56.2}  \\
\bottomrule[.9pt]
\end{tabular}
\vspace{-.4em}
\caption{\textbf{Ablation study about the multi-scale representation.} These results provide evidence for the benefits of employing a multi-scale representation in multimodal large language models.}
\label{tab:ab_multi}
}
\end{table*}

\begin{table*}[t]
\centering
{
\begin{tabular}{llcccccc}
\toprule[.9pt]
\textbf{POPE} &{\textbf{Methods}} & {\textbf{LLM Size}} & Accuracy & Precision & Recall & F1-Score & Yes \\ \midrule
 \multirow{2}{*}{{Random}} 
 & Single-scale & 7B & 73.88 & 67.03 & \bf{97.06} & 79.30 & 74.63 \\ 
 & \cellcolor{aliceblue!60} Multi-scale & \cellcolor{aliceblue!60} 7B & \cellcolor{aliceblue!60} \bf{85.19} & \cellcolor{aliceblue!60} \bf{83.59} & \cellcolor{aliceblue!60} 88.66 & \cellcolor{aliceblue!60} \bf{86.05} & \cellcolor{aliceblue!60} \bf{54.67}  \\ \midrule
  \multirow{2}{*}{{Popular}} 
 & Single-scale & 7B & 56.36 & 53.50 & \bf{97.20} & 69.01 & 90.83 \\ 
 & \cellcolor{aliceblue!60} Multi-scale & \cellcolor{aliceblue!60} 7B & \cellcolor{aliceblue!60} \bf{69.50} & \cellcolor{aliceblue!60} \bf{64.10} & \cellcolor{aliceblue!60} 88.60 & \cellcolor{aliceblue!60} \bf{74.39} & \cellcolor{aliceblue!60} \bf{69.10}  \\ \midrule
  \multirow{2}{*}{{Adversarial}} 
 & Single-scale & 7B & 55.63 & 53.07 & \bf{97.26} & 68.67 & 91.63 \\ 
 & \cellcolor{aliceblue!60} Multi-scale & \cellcolor{aliceblue!60} 7B & \cellcolor{aliceblue!60} \bf{64.97} & \cellcolor{aliceblue!60} \bf{60.23} & \cellcolor{aliceblue!60} 88.06 & \cellcolor{aliceblue!60} \bf{71.54} & \cellcolor{aliceblue!60} \bf{73.10}  \\ 
\bottomrule[.9pt]
\end{tabular}
\vspace{-.4em}
\caption{\textbf{Effect of the multi-scale representation on object hallucination.} ``Yes'' represents the proportion of positive answers that the model outputs.}
\label{tab:appendix_multi}
}
\end{table*}

\begin{table*}[t]
\centering
\footnotesize
{
\begin{tabular}{lcc|cccccc}
\toprule[1.pt]
\multirow{3}{*}{\textbf{Methods}} & \multicolumn{1}{c}{\textbf{Multimodal Pre-training}} & \multicolumn{1}{c|}{\textbf{Instruction Tuning}} & \multicolumn{4}{c}{\textbf{Image Understanding}} & \multirow{3}{*}{\textbf{POPE-R}} & \multirow{2}{*}{\textbf{Video}}\\
\cmidrule(rl){2-2}\cmidrule(rl){3-3}\cmidrule(rl){4-7}
 &{{Datasets}} &{{Datasets}} &{{Conv}} &{{Detail}} &{{Reason}} &{{All}} & & {\textbf{Inputs}} \\ \midrule
 LLaVA & \multirow{2}{*}{CC3M-595K} & \multirow{2}{*}{LLaVA-instruct-150K} & 82.3 & {70.2} & 87.9 & 80.4 & 66.83 & {\ding{56}} \\
 Chat-UniVi  & & & {82.9} & {68.8} & {89.8} & {80.7} & {82.26} & {\ding{56}}  \\ \midrule
 LLaVA & {CC3M-595K,} & \multirow{2}{*}{LLaVA-instruct-150K} & 82.7 & 68.8 & 88.8 & 80.8 & 72.02 & {\ding{56}} \\
 Chat-UniVi  & COCO & & {83.3} & {72.6} & {89.0} & {81.5} & {82.33} & {\ding{56}} \\ \midrule
 LLaVA & {CC3M-595K,} & {LLaVA-instruct-150K,} & 78.8 & {70.2} & {91.8} & 80.4 & 74.53 & {\ding{56}} \\
 Chat-UniVi  & COCO & MIMIC-IT-399K & {84.0} & 69.3 & 89.3 & {81.5} & {83.53} & {\ding{56}} \\ \midrule
 \rowcolor{aliceblue!60} Chat-UniVi & {CC3M-595K,} & {LLaVA-instruct-150K, MIMIC-IT-399K,} & \multirow{2}{*}{\Frst{84.1}} & \multirow{2}{*}{\Frst{74.2}} & \multirow{2}{*}{\Frst{93.7}} & \multirow{2}{*}{\Frst{84.2}} & \multirow{2}{*}{\Frst{85.19}} & \multirow{2}{*}{\ding{52}} \\
  w/ video data & COCO & Video-ChatGPT-instruct & \\ 
\bottomrule[1.pt]
\end{tabular}
\vspace{-.4em}
\caption{\textbf{Ablation of structure and training data.} ``\ding{56}'' denotes that the method does not have this property. ``\ding{52}'' denotes that the method has this property.}
\label{tab:ablation of data}
}
\end{table*}

\section{Additional Experiments}\label{appendix:Additional Experiments}
\noindent \myparagraph{Comparison between the LoRA and Full Fine-tuning.}
When the number of model parameters is too large, full fine-tuning of retraining all model parameters becomes expensive, so many recent methods freeze most of the model parameters and train the model with LoRA~\cite{hu2022lora}. We provide the results of the comparison between the LoRA and full fine-tuning in Tab.~\ref{tab:appendix_lora}. We find that LoRA can achieve competitive performance with full fine-tuning while saving more than half the GPU memory required for training. Future work can use LoRA to extend our method on larger LLMs and vision encoders to achieve better performance.

\noindent \myparagraph{Analysis of the Vision Encoder.}
EVA-CLIP~\cite{sun2023eva} is a recently developed multimodal model with performance comparable to Openai-CLIP~\cite{radford2021learning}. We provide the results of the comparison between EVA-CLIP and Openai-CLIP in Tab.~\ref{tab:appendix_eva}. We find that the performance of EVA-CLIP is comparable to that of Openai-CLIP when the number of parameters is equal. However, EVA-CLIP offers a larger version of the model with a parameter count of 1.8B, so we think it might be better to adopt a larger EVA-CLIP than Openai-CLIP when using larger LLMs.

\begin{table*}[t]
\centering
{
\begin{tabular}{llcccccc}
\toprule[.9pt]
\textbf{POPE} &{\textbf{Methods}} & {\textbf{LLM Size}} & Accuracy & Precision & Recall & F1-Score & Yes \\ \midrule
 \multirow{7}{*}{{Random}} & \largemodel LLaVA & \largemodel 13B & \largemodel 64.12 & \largemodel 59.38 & \largemodel 95.99 & \largemodel 73.38 & \largemodel 83.26 \\
 & \largemodel MiniGPT-4 & \largemodel 13B & \largemodel 79.67 & \largemodel 78.24 & \largemodel 82.20 & \largemodel 80.17 & \largemodel 52.53 \\ 
 & \largemodel InstructBLIP & \largemodel 13B & \largemodel 88.57 & \largemodel 84.09 & \largemodel 95.13 & \largemodel 89.27 & \largemodel 56.57  \\ 
 & MultiModal-GPT & 7B & 50.10 & 50.05 & \bf{100.00} & 66.71 & 99.90\\
 & mPLUG-Owl & 7B & 53.97 & 52.07 & 99.60 & 68.39 & 95.63  \\
 & LLaVA\ssymbol{2} & 7B & 72.16 & 78.22 & 76.29 & 78.22 & 76.29 \\ 
 & \cellcolor{aliceblue!60} Chat-UniVi & \cellcolor{aliceblue!60} 7B & \cellcolor{aliceblue!60} \bf{85.19} & \cellcolor{aliceblue!60} \bf{83.59} & \cellcolor{aliceblue!60} 88.66 & \cellcolor{aliceblue!60} \bf{86.05} & \cellcolor{aliceblue!60} \bf{54.67}  \\ \midrule
  \multirow{7}{*}{{Popular}} & \largemodel LLaVA & \largemodel 13B & \largemodel 63.90 & \largemodel 58.46 & \largemodel 95.86 & \largemodel 72.63 & \largemodel 81.93 \\
 & \largemodel MiniGPT-4 & \largemodel 13B & \largemodel 69.73 & \largemodel 65.86 & \largemodel 81.93 & \largemodel 73.02 & \largemodel 62.20 \\ 
 & \largemodel InstructBLIP & \largemodel 13B & \largemodel 82.77 & \largemodel 76.27 & \largemodel 95.13 & \largemodel 84.66 & \largemodel 62.37  \\ 
 & MultiModal-GPT & 7B & 50.00 & 50.00 & \bf{100.00} & 66.67 & 100.00\\
 & mPLUG-Owl & 7B & 50.90 & 50.46 & 99.40 & 66.94 & 98.57  \\
 & LLaVA\ssymbol{2} & 7B & 61.37 & 56.63 & 97.00 & 71.52 & 85.63 \\ 
 & \cellcolor{aliceblue!60} Chat-UniVi & \cellcolor{aliceblue!60} 7B & \cellcolor{aliceblue!60} \bf{69.50} & \cellcolor{aliceblue!60} \bf{64.10} & \cellcolor{aliceblue!60} 88.60 & \cellcolor{aliceblue!60} \bf{74.39} & \cellcolor{aliceblue!60} \bf{69.10}  \\ \midrule
  \multirow{7}{*}{{Adversarial}} & \largemodel LLaVA & \largemodel 13B & \largemodel 58.91 & \largemodel 55.11 & \largemodel 95.72 & \largemodel 69.95 & \largemodel 86.76 \\
 & \largemodel MiniGPT-4 & \largemodel 13B & \largemodel 65.17 & \largemodel 61.19 & \largemodel 82.93 & \largemodel 70.42 & \largemodel 67.77 \\ 
 & \largemodel InstructBLIP & \largemodel 13B & \largemodel 72.10 & \largemodel 65.13 & \largemodel 95.13 & \largemodel 77.32 & \largemodel 73.03  \\ 
 & MultiModal-GPT & 7B & 50.00 & 50.00 & \bf{100.00} & 66.67 & 100.00\\
 & mPLUG-Owl & 7B & 50.67 & 50.34 & 99.33 & 66.82 & 98.67  \\
 & LLaVA\ssymbol{2} & 7B & 58.67 & 54.90 & 97.00 & 70.12 & 88.33 \\ 
 & \cellcolor{aliceblue!60} Chat-UniVi & \cellcolor{aliceblue!60} 7B & \cellcolor{aliceblue!60} \bf{64.97} & \cellcolor{aliceblue!60} \bf{60.23} & \cellcolor{aliceblue!60} 88.06 & \cellcolor{aliceblue!60} \bf{71.54} & \cellcolor{aliceblue!60} \bf{73.10}  \\ 
\bottomrule[.9pt]
\end{tabular}
\vspace{-.4em}
\caption{\textbf{Detailed results on object hallucination evaluation.} ``\ssymbol{2}'' denotes our own re-implementation of LLaVA under our training settings (excluding video data) for a fair comparison.}
\label{tab:appendix_hallucination}
}
\end{table*}

\noindent \myparagraph{Effect of the Multi-scale Representation.} 
To investigate the impact of the multi-scale representation of our method, we provide the ablation results in Tab.~\ref{tab:ab_multi}. Multi-scale representation improves both image understanding and video understanding of the model. These results provide evidence for the benefits of employing a multi-scale representation in multimodal large language models.

\noindent \myparagraph{Effect of the Multi-scale Representation on Object Hallucination.}
As shown in Tab.~\ref{tab:pope}, Chat-UniVi, as a 7B model, even outperforms the 13B model, \eg, MiniGPT-4, in the object hallucination evaluation. We attribute this success to the multi-scale representation that equips our method to perceive both high-level semantic concepts and low-level visual appearance. In Tab.~\ref{tab:appendix_multi}, we show the results of ablation experiments on object hallucination evaluation for the multi-scale representation. We find that multi-scale representation improves the ability to resist hallucinations. Therefore, multi-scale representation is beneficial for multimodal LLMs.

\noindent \myparagraph{Ablation of Training Data.}
We provide comparisons of our method with LLaVA under different conditions in Tab.~\ref{tab:ablation of data}. Our method achieves better performance than LLaVA, which we explain in the following two aspects. \textbf{{Multi-scale Representation.}} In contrast to LLaVA, which focuses on low-level visual features, our method perceives both high-level semantic concepts and low-level visual details by multi-scale representation. Therefore, our method outperforms LLaVA in conversation, reasoning, and hallucinations. \textbf{{Scalability.}} Our framework supports video input, and by fine-tuning with high-quality video instruction data, the visual capabilities of our models have been significantly enhanced, especially in terms of detailed captioning and reasoning.

Besides, we draw the following two conclusions: (1) Instruction tuning data has a greater impact on performance than pre-training data. (2) High-quality instruction tuning data can significantly enhance model performance. Especially after training on high-quality video data, the performance of the model is greatly improved.

\noindent \myparagraph{Detailed Results on Object Hallucination Evaluation.}
In Tab.~\ref{tab:appendix_hallucination}, we report the detailed results of the polling-based object probing evaluation~\cite{li2023evaluating}. As shown in Tab.~\ref{tab:appendix_hallucination}, Chat-UniVi outperforms the recently proposed state-of-the-art methods. Notably, as a 7B model, our method even outperforms the 13B model, \eg, MiniGPT-4, in the object hallucination evaluation. These results demonstrate the effectiveness of our method.

\begin{figure*}[t]
\centering
\includegraphics[width=1\textwidth]{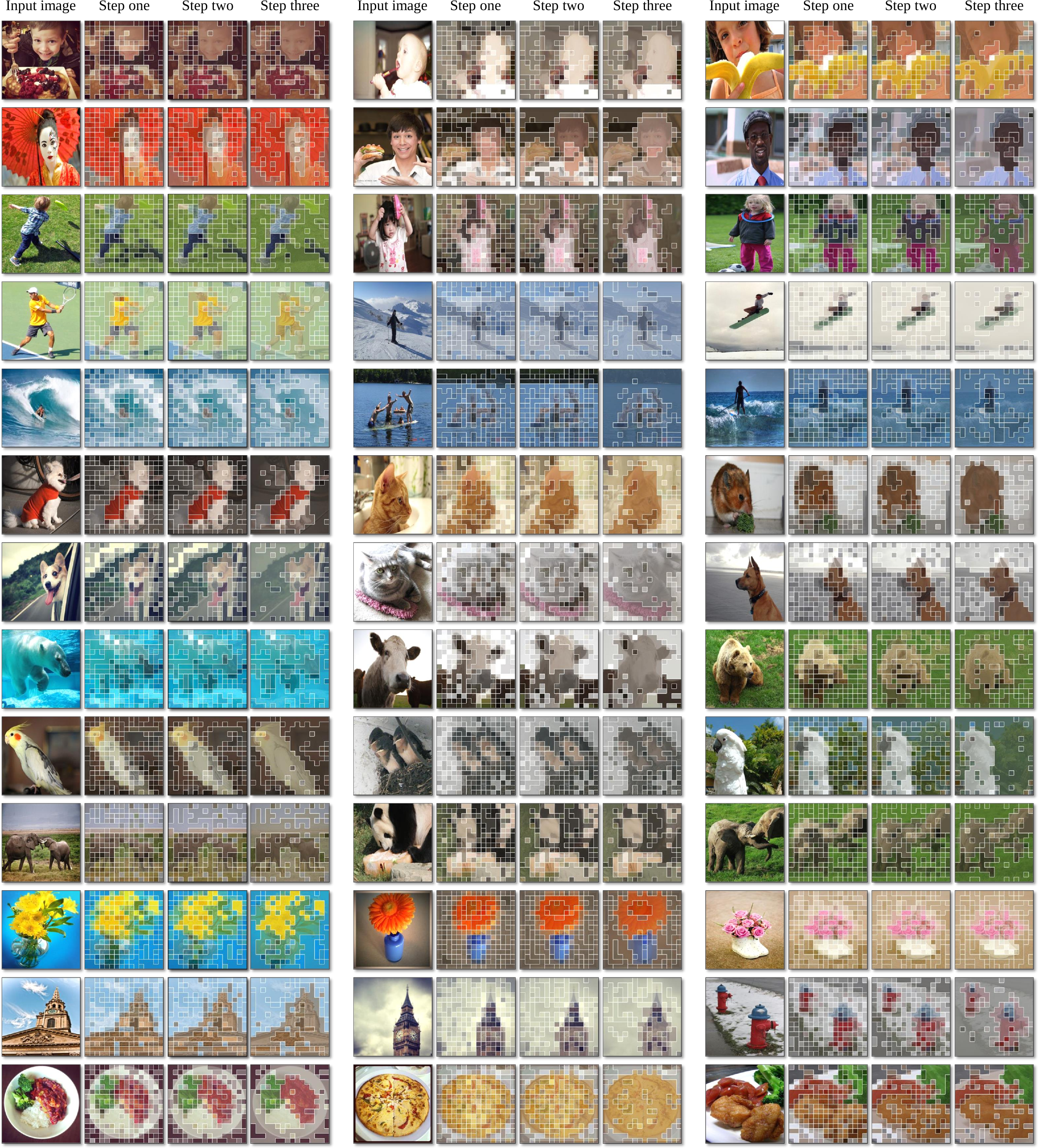}
\caption{\textbf{Visualization of the dynamic visual tokens for the image inputs.} We provide a diverse range of visualizations encompassing various image categories, including portraits, sports, wildlife, art, architecture, and food. It is important to emphasize that our proposed token merging method is parameter-free and operates without the need for object outline labels.}
\vspace{-1.em}
\label{fig:appendix_dynamic visual tokens0}
\end{figure*}

\begin{figure*}[t]
\centering
\includegraphics[width=1\textwidth]{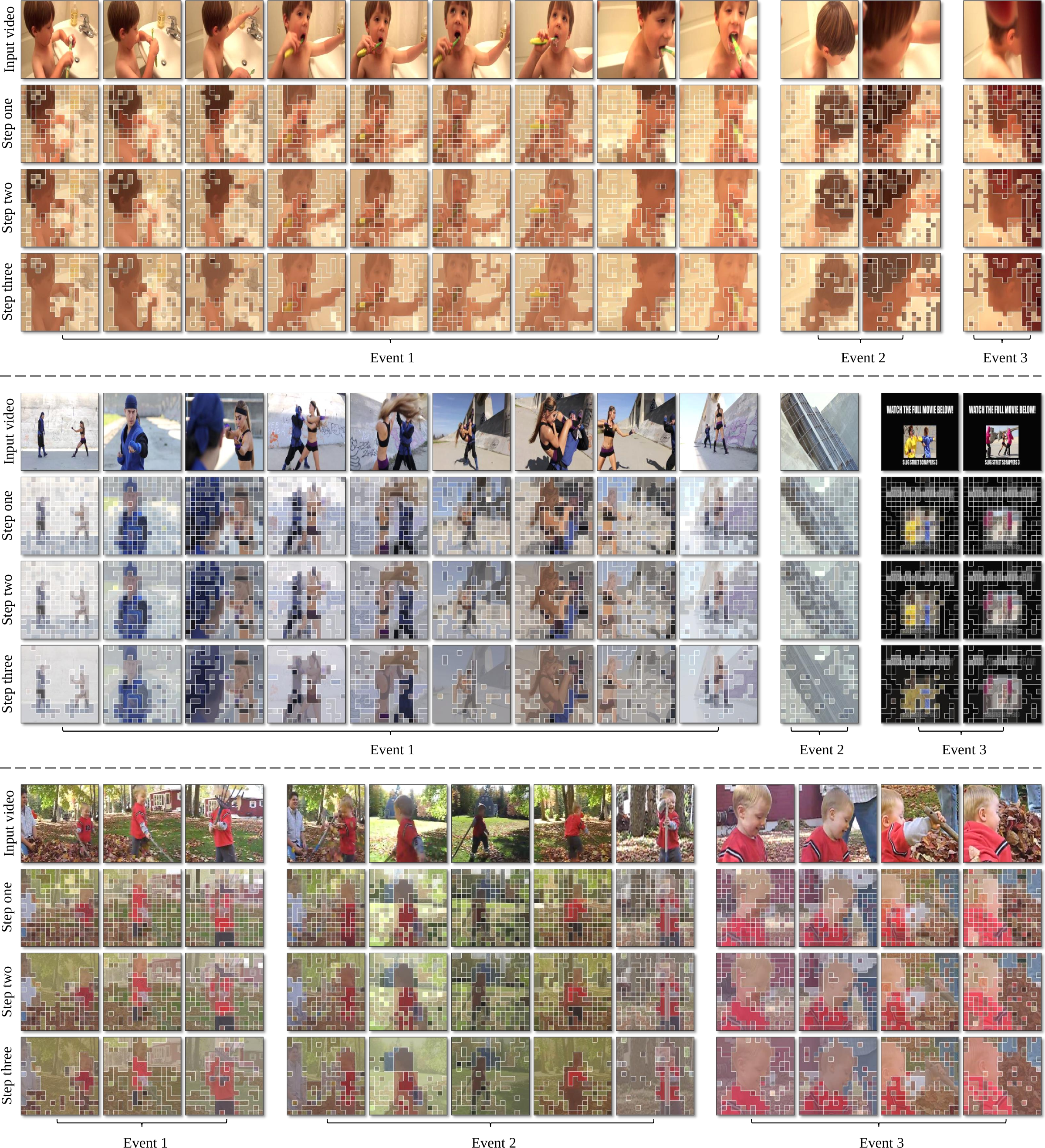}
\caption{\textbf{Visualization of the dynamic visual tokens for the video inputs.} It is important to emphasize that our proposed token merging method is parameter-free and operates without the need for object outline labels. Our method imposes no restrictions on the number of frames per event, showcasing the remarkable flexibility and generalization ability of our methodology.}
\vspace{-1.em}
\label{fig:appendix_dynamic visual tokens1}
\end{figure*}

\section{Additional Visualization Results}\label{appendix:Additional Visualization Results}
\noindent \myparagraph{Visualization of the dynamic visual tokens for the image inputs.} 
To gain a deeper insight into the functionality of our proposed dynamic visual tokens, we present the additional visualization results for the image inputs in Fig.~\ref{fig:appendix_dynamic visual tokens0}. In Fig.~\ref{fig:appendix_dynamic visual tokens0}, we provide a diverse range of visualizations encompassing various image categories, including portraits, sports, wildlife, art, architecture, and food. It is crucial to underscore that our proposed token merging method operates without the need for object outline labels and is parameter-free. As shown in Fig.~\ref{fig:appendix_dynamic visual tokens0}, the proposed dynamic visual tokens effectively generalize objects and backgrounds, empowering Chat-UniVi to capture the spatial nuances of images using a limited number of visual tokens.

\noindent \myparagraph{Visualization of the dynamic visual tokens for the video inputs.} 
To gain a more comprehensive understanding of our proposed dynamic visual tokens, we also present additional visualization results for the video inputs in Fig.~\ref{fig:appendix_dynamic visual tokens1}. In the case of videos, the video is initially divided into several events, and subsequently, these visual tokens expand over frames within each event to encapsulate frame-level dynamics. Notably, our method imposes no restrictions on the number of frames per event, showcasing the remarkable flexibility and generalization ability of our methodology. As shown in Fig.~\ref{fig:appendix_dynamic visual tokens1}, the proposed dynamic visual tokens significantly reduce the number of visual tokens while maintaining the expressive capabilities of the model. This empowerment equips Chat-UniVi with the capacity to capture the broader temporal understanding required for videos, all within the confines of a limited number of visual tokens.

\section{Additional Qualitative Analysis}\label{appendix:Additional Qualitative Analysis}
\begin{figure*}[tbp]
\centering
\includegraphics[width=0.8\textwidth]{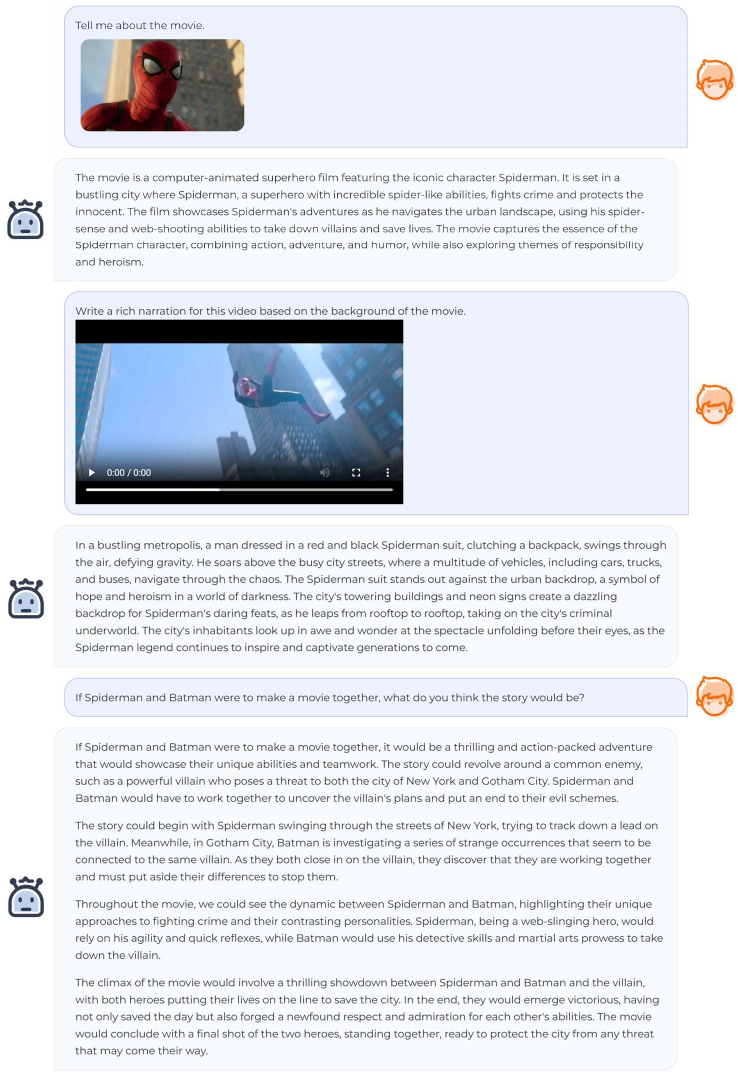}
\caption{\textbf{A conversation with both image and video.} The blue box shows the user input. The gray box shows the model output.}
\label{fig:appendix_demo0}
\end{figure*}

\begin{figure*}[tbp]
\centering
\includegraphics[width=0.8\textwidth]{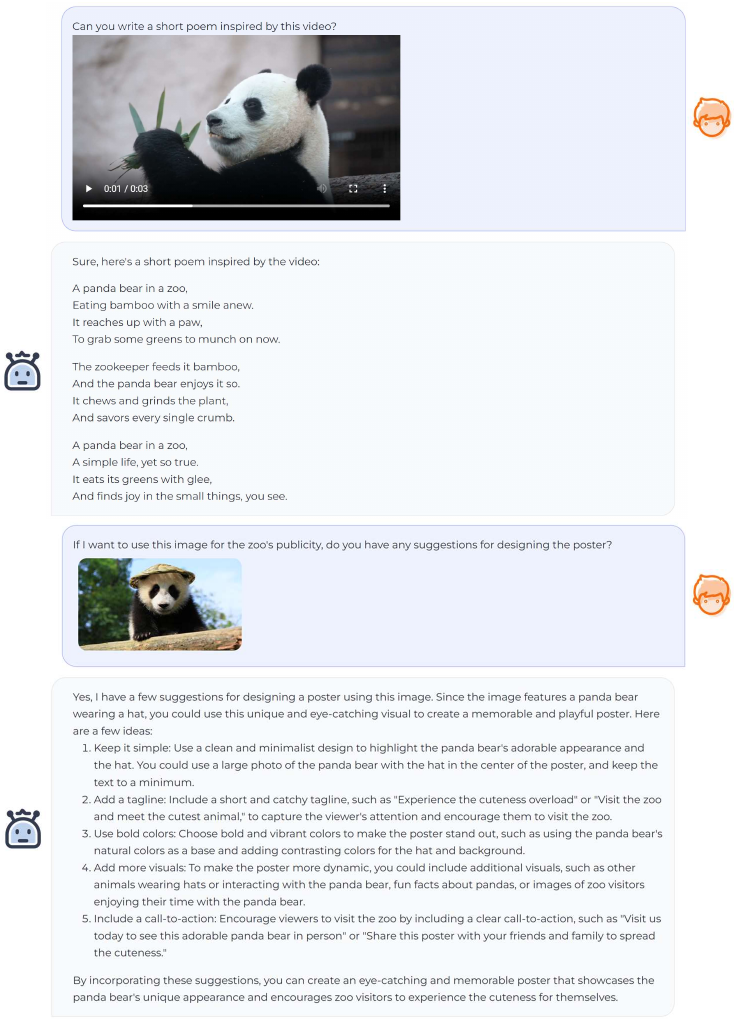}
\caption{\textbf{A conversation with both image and video.} The blue box shows the user input. The gray box shows the model output.}
\label{fig:appendix_demo1}
\end{figure*}

\noindent \myparagraph{The conversation includes both the image and the video.} 
In Fig.~\ref{fig:appendix_demo0} and Fig.~\ref{fig:appendix_demo1}, we present examples of conversations that encompass both the image and the video. As shown in Fig.~\ref{fig:appendix_demo0} and Fig.~\ref{fig:appendix_demo1}, Chat-UniVi offers detailed and contextually appropriate responses aligned with user prompts. These illustrative examples showcase the remarkable ability of Chat-UniVi to comprehend both image and video contexts across multiple conversational turns.

\begin{figure*}[tbp]
\centering
\includegraphics[width=0.8\textwidth]{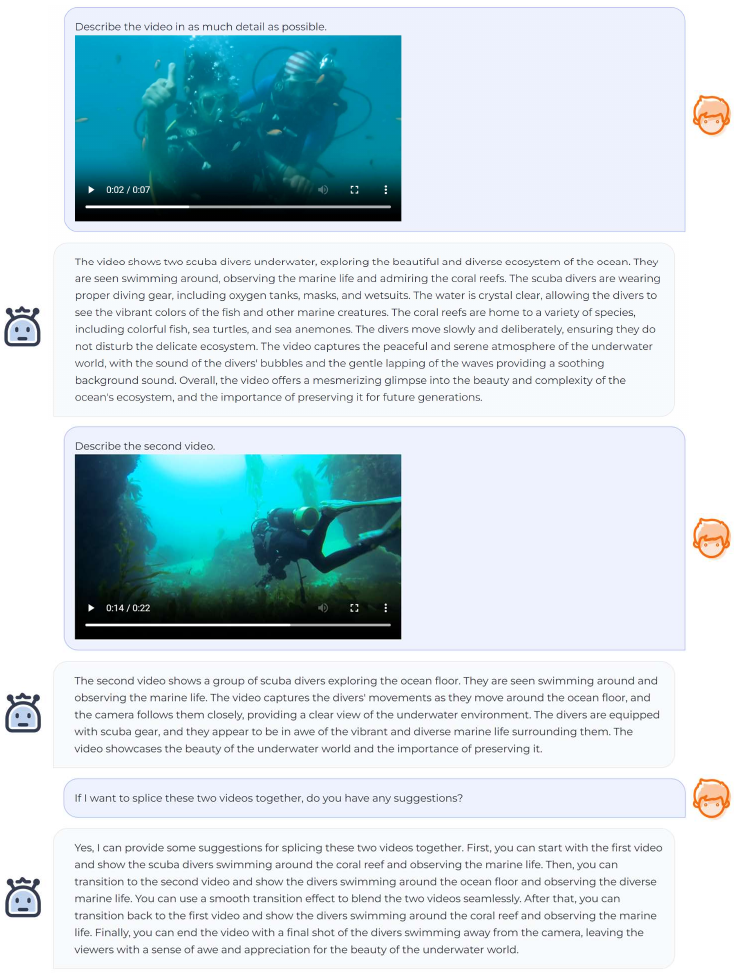}
\caption{\textbf{A conversation includes multiple videos.} The blue box shows the user input. The gray box shows the model output.}
\label{fig:appendix_demo2}
\end{figure*}

\noindent \myparagraph{The conversation includes multiple videos.} 
Fig.~\ref{fig:appendix_demo2} illustrates a conversation example including multiple videos. As shown in Fig.~\ref{fig:appendix_demo2}, Chat-UniVi can use the information of multiple videos in the context, and provide appropriate and coherent responses based on user prompts. The illustrative example showcases the remarkable ability of Chat-UniVi to comprehend multiple video contexts across multiple conversational turns.

\begin{figure*}[tbp]
\centering
\includegraphics[width=0.8\textwidth]{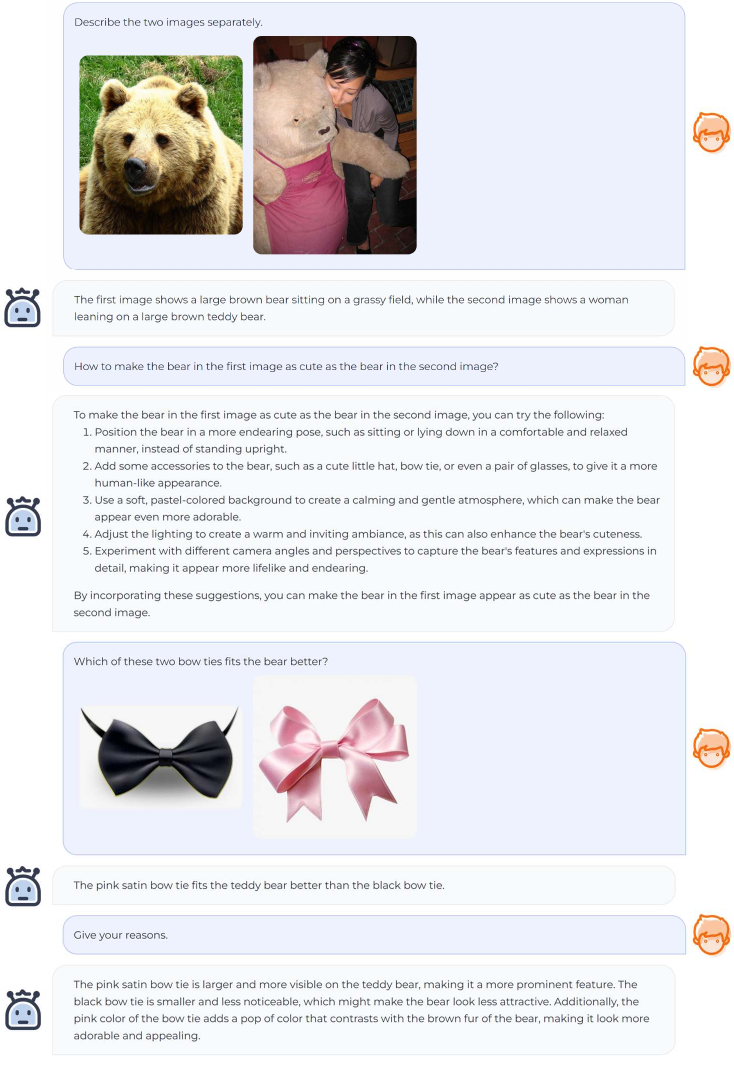}
\caption{\textbf{A conversation includes multiple images.} The blue box shows the user input. The gray box shows the model output.}
\label{fig:appendix_demo3}
\end{figure*}

\noindent \myparagraph{The conversation includes multiple images.} 
Fig.~\ref{fig:appendix_demo3} provides an illustrative conversation example including multiple images. As shown in Fig.~\ref{fig:appendix_demo3}, Chat-UniVi adeptly leverages information from multiple images within the context, enabling it to make choices among various images. This illustrative example highlights the impressive capacity of Chat-UniVi to grasp multiple image contexts seamlessly throughout various conversational exchanges.

\begin{figure*}[tbp]
\centering
\includegraphics[width=0.8\textwidth]{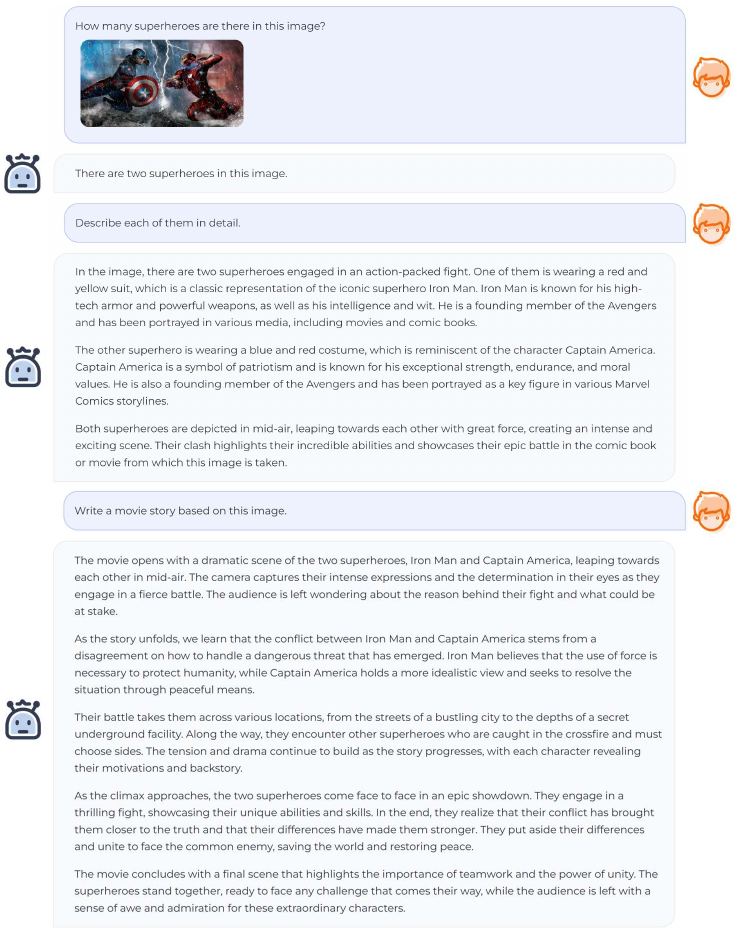}
\caption{\textbf{A conversation includes the image.} The blue box shows the user input. The gray box shows the model output.}
\label{fig:appendix_demo4}
\end{figure*}

\noindent \myparagraph{The conversation includes the image.} 
Fig.~\ref{fig:appendix_demo4} features an example of a conversation that incorporates an image. As shown in Fig.~\ref{fig:appendix_demo4}, Chat-UniVi excels at providing detailed descriptions and can even craft compelling narratives inspired by the image. The illustrative example showcases the remarkable ability of Chat-UniVi in the realms of reasoning and creative expression.

\begin{figure*}[tbp]
\centering
\includegraphics[width=0.8\textwidth]{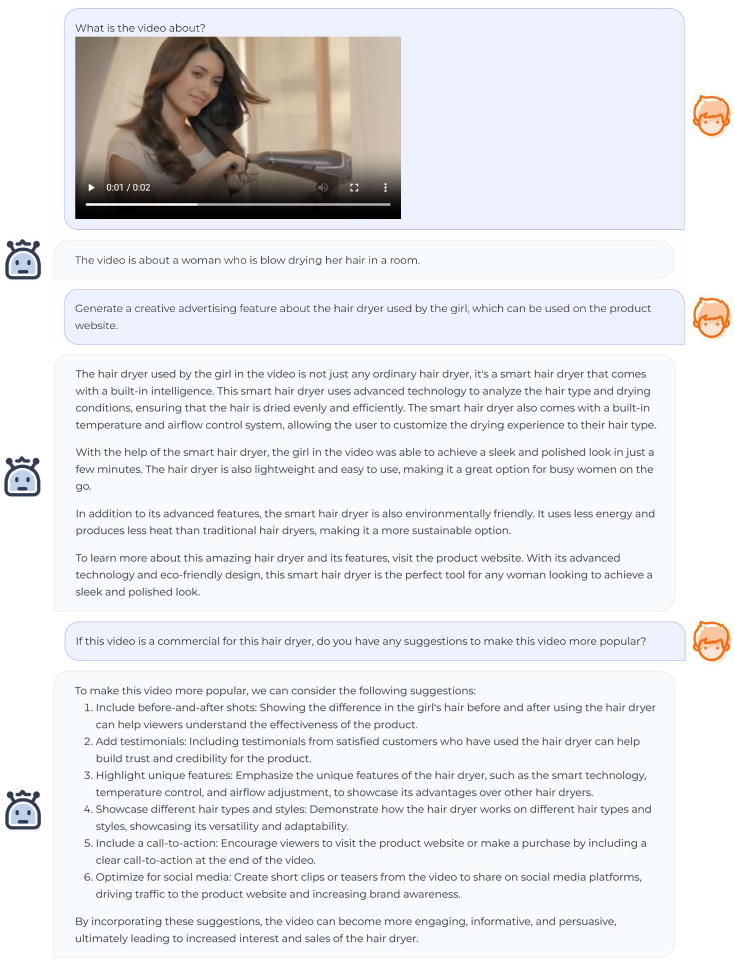}
\caption{\textbf{A conversation includes the video.} The blue box shows the user input. The gray box shows the model output.}
\label{fig:appendix_demo5}
\end{figure*}

\begin{figure*}[tbp]
\centering
\includegraphics[width=0.8\textwidth]{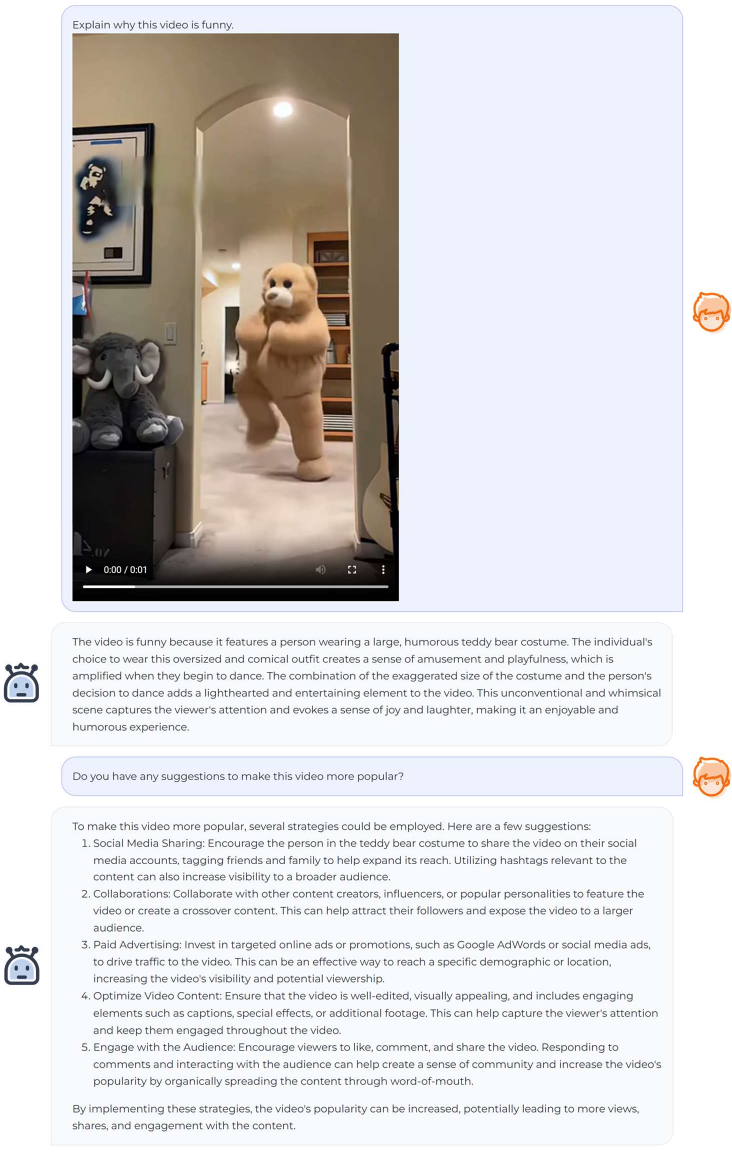}
\caption{\textbf{A conversation includes the video.} The blue box shows the user input. The gray box shows the model output.}
\label{fig:appendix_demo6}
\end{figure*}

\noindent \myparagraph{The conversation includes the video.} 
In Fig.~\ref{fig:appendix_demo5} and Fig.~\ref{fig:appendix_demo6}, we offer examples of conversations that incorporate the video. As shown in Fig.~\ref{fig:appendix_demo5} and Fig.~\ref{fig:appendix_demo6}, Chat-UniVi exhibits a remarkable proficiency in comprehending videos and is adept at offering valuable insights inspired by the video content. These illustrative examples showcase the remarkable ability of Chat-UniVi to grasp video contexts and engage in reasoned responses.

\section{Details of Quantitative Evaluations}\label{appendix:Details of Quantitative Evaluations}
\noindent \myparagraph{GPT-based Evaluation For Image Understanding.}
Our quantitative evaluation protocol follows that of \citet{liu2023visual}. Following \citet{liu2023visual,zhang2023llavar}, we employ 90 questions based on 30 COCO validation images, covering various aspects, including conversation, detail description (Detail), and complex reasoning (Reason). These images are randomly selected by \citet{liu2023visual}. We utilize the GPT-4 model to generate reference responses based on the question, and the ground-truth bounding boxes and captions. During the model evaluation process, the model predicts answers based on both the question and input image. After obtaining the response from the model, we feed the question, visual information (in the format of captions and bounding boxes), the generated response, and the reference response to GPT-4. GPT-4 evaluates the helpfulness, relevance, accuracy, and level of detail of the responses, assigning an overall score on a scale of 1 to 10, where a higher score indicates better overall performance. Besides, we also ask GPT-4 to provide a comprehensive explanation of the evaluation to enhance our understanding of the models.

\noindent \myparagraph{GPT-based Evaluation For Video Understanding.}
The quantitative evaluation protocol for video understanding follows the methodology introduced by \citet{maaz2023video}. Specifically, \citet{maaz2023video} curates a test set based on the ActivityNet-200 dataset~\cite{caba2015activitynet}, which includes videos with rich, dense descriptive captions and associated question-answer pairs from human annotations. During the model evaluation process, we employ the GPT-3.5 model to assign a relative score to the generated predictions on a scale of 1-5, across five critical aspects: (1) Correctness of information (Correct). (2) Detail orientation (Detail). (3) Contextual understanding (Context). (4) Temporal understanding (Temporal). (5) Consistency. It is worth noting that the results reported in \citet{maaz2023video} span a range from 0 to 5. To standardize the metrics, we normalize all scores to a scale of 0 to 100.

\noindent \myparagraph{Zero-shot Video Question Evaluation.}
Our evaluation protocol follows that of \citet{maaz2023video}, utilizing GPT-assisted evaluation to assess the capabilities of models. During the model evaluation process, we feed the question, the ground-truth answer, and the generated response to the GPT-3.5 model. GPT-3.5 evaluates whether the generated responses are correct and assigns a matching score on a scale of 0 to 5, where a higher score indicates better overall performance.

\noindent \myparagraph{Zero-shot Object Hallucination Evaluation.}
To quantitatively evaluate the hallucination problem of the model, we adopt the polling-based object probing evaluation (POPE) process proposed by \citet{li2023evaluating}. Specifically, POPE formulates the evaluation of object hallucination as a binary classification task, where the model is prompted to respond with either ``Yes'' or ``No'' to queries like ``Is there a chair in the image?''. \citet{li2023evaluating} randomly selects 500 images from the COCO validation set. Each image contains more than three ground-truth objects in the annotations, and six questions are generated for each image. The annotations of objects in images directly construct the questions with the answer ``Yes''. For the questions with the answer ``No'', three different strategies are employed for sampling their probing objects as follows:

\begin{itemize}
\item \textbf{Random Sampling.} Randomly sampling objects that do not exist in the image.

\item \textbf{Popular Sampling.} Selecting the top-3 most frequently occurring objects in the COCO dataset that are absent from the image.

\item \textbf{Adversarial Sampling.} Initially, \citet{li2023evaluating} rank all objects based on their co-occurring frequencies with the ground-truth objects, and subsequently select the top-3 most frequent objects from this list that are not present in the image.

\end{itemize}

\end{document}